\documentclass{IEEEtran}
\IEEEoverridecommandlockouts
\usepackage{amsmath,amssymb,amsfonts}
\usepackage{textcomp}
\def\BibTeX{{\rm B\kern-.05em{\sc i\kern-.025em b}\kern-.08em
		T\kern-.1667em\lower.7ex\hbox{E}\kern-.125emX}}

\usepackage{marvosym}
\usepackage{bm}
\usepackage{diagbox}

\usepackage{booktabs}
\usepackage{colortbl}
\usepackage[pdftex]{graphicx}
\graphicspath{{../pdf/}}
\usepackage{amsmath}
\usepackage{algorithm}
\usepackage{algorithmic}
\usepackage{array}
\usepackage{multirow}
\usepackage[numbers,compress]{natbib}

\usepackage{graphicx}
\ifCLASSOPTIONcompsoc
\usepackage[caption=false, font=normalsize, labelfont=sf, textfont=sf]{subfig}
\else
\usepackage[caption=false, font=footnotesize]{subfig}
\fi
\usepackage{url}
\usepackage{color,xcolor}

\newcommand{\fref}[1]{Fig. \ref{#1}}

\makeatletter

\usepackage{amsfonts,dsfont}
\usepackage{hyperref} 
\usepackage{multirow}
\usepackage{threeparttable}

\begin{document}
	
	\title{Meta-Learning-Based Deep Reinforcement Learning for Multiobjective Optimization Problems}
	
	\author{Zizhen Zhang,~\IEEEmembership{Member,~IEEE}, Zhiyuan Wu, Hang Zhang, and Jiahai Wang$^*$,~\IEEEmembership{Senior Member,~IEEE}
		
		\thanks{
			Zizhen Zhang, Zhiyuan Wu, Hang Zhang and Jiahai Wang are with the School of Computer Science and Engineering, Sun Yat-sen University, Guangzhou 510275, China, and also with the Guangdong Key Laboratory of Big Data Analysis and Processing, Sun Yat-sen University, Guangzhou 510275, China (\textit{e-mail}: \href{mailto:zhangzzh7@mail.sysu.edu.edu}{zhangzzh7@mail.sysu.edu.cn}; \href{mailto:wuzhy66@mail2.sysu.edu.cn}{wuzhy66@mail2.sysu.edu.cn};
			\href{mailto:zhangh669@mail2.sysu.edu.cn}{zhangh669@mail2.sysu.edu.cn};
			\href{mailto:wangjiah@mail.sysu.edu.cn}{wangjiah@mail.sysu.edu.cn}).

			
		}
	}
	
	\maketitle
	\begin{abstract}
		Deep reinforcement learning (DRL) has recently shown its success in tackling complex combinatorial optimization problems. When these problems are extended to multiobjective ones, it becomes difficult for the existing DRL approaches to flexibly and efficiently deal with multiple subproblems determined by weight decomposition of objectives. This paper proposes a concise meta-learning-based DRL approach. It first trains a meta-model by meta-learning. The meta-model is fine-tuned with a few update steps to derive submodels for the corresponding subproblems. The Pareto front is then built accordingly. Compared with other learning-based methods, our method can greatly shorten the training time of multiple submodels. Due to the rapid and excellent adaptability of the meta-model, more submodels can be derived so as to increase the quality and diversity of the found solutions. The computational experiments on multiobjective traveling salesman problems and multiobjective vehicle routing problem with time windows demonstrate the superiority of our method over most of learning-based and iteration-based approaches.
	\end{abstract}
	
	
	\section{Introduction}
	\label{Intro}
	A multiobjective optimization problem (MOP) can be generally defined as follows: 
	\begin{equation}
		\begin{array}{ll}{\operatorname{minimize}} & {f(x)=\left(f_{1}(x), f_{2}(x), \ldots, f_{m}(x)\right)} \\ {\text { s.t. }} & {x \in X,}
		\end{array}
		\label{eq:mop}
	\end{equation}
	where $f(x)$ consists of $m$ different objective functions and $X \subseteq \mathbb{R}^n$ is the decision space. Since these $m$ objectives are usually conflicting with each other, a set of trade-off solutions, termed Pareto optimal solutions, are sought for MOPs. Formally, for two objectives $u, v \in \mathbb{R}^m$, $u$ is said to dominate $v$, if and only if $u_i\leq v_i$ for every $i \in \{1,2,\dots,m \}$ and $u_j < v_j$ for at least one $j \in \{1,2,\dots,m \}$. A solution $x^* \in X$ is called a Pareto optimal solution, if there is no solution $x \in X$ such that $f(x)$ dominates $f(x^*$). All the Pareto optimal solutions constitute a Pareto set (PS). The corresponding objective vectors $\{f(x^*)| x^* \in \mbox{PS}\}$ constitute a Pareto front (PF). 
	
	In order to solve MOPs, two paradigms are commonly considered: a priori methods and posteriori methods. For a priori methods, one asks for the preference of a decision maker (DM) towards different objectives, and finds a preferred Pareto optimal solution at a time. For posteriori methods, on the other hand, one generates a group of Pareto optimal solutions and asks DM to select the most preferred one. The drawback of the former paradigm mainly lies in that DM is difficult to express the preferences before seeing any Pareto optimal solutions. In this case, DM may frequently post queries to look for a desirable solution, thereby leading to large computation time. The drawback of the latter paradigm would be that approximating the PF often requires a number of iterations and thus is time consuming. In summary, both paradigms indicate the computational difficulty in tackling MOPs.
	
	In literature, MOPs are widely studied by computational intelligence communities. With regard to a priori methods, they essentially treat the MOP as a single-objective optimization problem. Various kinds of optimization approaches, mostly heuristics, can be adopted \cite{gendreau2010handbook}. To speed up the search of goals, it is quite natural to apply the idea of machine learning. By using advanced machine learning techniques, a high-quality solution can be efficiently generated once a model has been well trained. In particular, with the recent development of deep learning (DL), more and more DL models have shown their capability in tackling complex optimization tasks. In more recent years, deep reinforcement learning (DRL) has been proposed in the field of combinatorial optimization \cite{bengio2020machine,vesselinova2020learning,mazyavkina2020reinforcement}, which indicates the potential of leveraging learning-based approaches to replace traditional exact or heuristic algorithms for solving combinatorial optimization problems \cite{bello2017neural,nazari2018reinforcement,kool2019attention}. 
	
	With regard to posteriori methods, they are extensively investigated in the scope of evolutionary computation. Some classic multiobjective evolutionary algorithms (MOEAs), such as NSGA-$\rm\uppercase\expandafter{\romannumeral2}$ \cite{deb2002fast} and MOEA/D \cite{zhang2007moea}, have been demonstrated their great success in dealing with many practical applications of MOPs. In addition, some handcrafted heuristics are integrated into MOEAs to further enhance the search quality. Typical approaches include Pareto local search (PLS) \cite{angel2004dynasearch} and multiple objective genetic local search algorithm (MOGLS) \cite{jaszkiewicz2002genetic}.
	
	No matter a priori or posteriori methods, the existing DRL methods for MOPs require predetermined objective weight vectors before training the models. Hence, the quality of the obtained solution may depend heavily on whether the weight vector has been observed and trained by DRL. If an unseen weight vector is encountered, a new model needs to be trained from scratch, or from model transferring, for the corresponding subproblem. For practical usage, a large number of models need to be pretrained and stored to meet the potential subproblems.
	
	In fact, training a specific model for a specific weight vector (or subproblem) is ineffective. We therefore introduce the idea of meta-learning, and train a meta-model to handle all potential subproblems. Meta-learning, also known as learning to learn, is the science of systematically observing how different machine learning approaches perform on a wide range of learning tasks, and then learning from the experience to learn new tasks much faster \cite{vanschoren2018meta}. Our work is originally inspired by several recently proposed meta-learning algorithms for fast adaptation of deep networks \cite{finn2017model,nichol2018first}. Once a meta-model is trained, given any weight vector, the meta-model can quickly adapt to a submodel with a few fine-tuning steps for the corresponding subproblem.
	
	In this paper, we propose a meta-learning-based DRL (MLDRL) to tackle MOPs. Two classic MOPs, i.e., multiobjective traveling salesman problem (MOTSP) and multiobjective vehicle routing problem with time windows (MOVRPTW), are chosen as benchmark problems to demonstrate the effectiveness of the proposed method. Both problems are sufficiently challenging and have many practical applications. Our main contributions can be highlighted as follows.
	\begin{itemize}
		\item Our work provides a new, flexible and efficient way of solving MOPs by means of MLDRL. To our best knowledge, this is the first time that MLDRL has been introduced in the field of multiobjective optimization, especially for complex combinatorial problems.
		\item The proposed MLDRL is applicable in either a priori or posteriori schemes. Compared with previous learning-based methods, it can greatly shorten the training time of multiple submodels. When new subproblems (with unseen weight vectors) appear, we only need to fine-tune the meta-model with a few gradient updates to obtain satisfactory solutions. When constructing the PF, due to the rapid and excellent adaptability of the meta-model, we can easily fine-tune the meta-model for more weight vectors to increase the quality and diversity of the found solutions. 
		\item Compared with the classic iteration-based approaches, MLDRL can achieve better performance in a short time. In addition, MLDRL has better generalization ability than previous learning-based methods. Once a meta-model is trained, not only the submodels fine-tuned from the meta-model, but also the meta-model itself can be well generalized to the problem instances with different scales.
	\end{itemize}
	
	The remainder of this paper is organized as follows. Section \ref{r} reviews the related work of MLDRL. Section \ref{2} introduces common techniques for MOPs. Section \ref{3} presents the proposed MLDRL. Section \ref{4} discusses the deep learning models for two benchmark problems. Section \ref{5} and Section \ref{exp:mocvrptw} provide the computational results with detailed analyses. Section \ref{6} concludes our work.
	
	\section{Related Work}
	\label{r}
	In literature, the learning-based methods for complex combinatorial optimization problems can be roughly divided into a group of end-to-end DRL approaches and a group of heuristic search approaches guided by DRL. Because our study focuses on the classic TSP and VRPTW, we classify the related literature into these two groups and review some representative works as follows.
	
	End-to-end DRL is to train a deep neural network to directly learn how to make sequential decisions from raw data. It can be regarded as construction-based. Starting from an empty solution, a decision is made at each step until a complete solution is obtained. \citet{bello2017neural} first used a DRL approach to solve TSP, in which a deep model called pointer network \cite{pointer}, is adopted. \citet{dai2017learning} designed a graph embedding network to extract features of the graph and applied a Q-learning algorithm for the training. A solution is constructed incrementally according to the output of a graph embedding network. They evaluated their method on minimum vertex cover problem, maximum cut problem and TSP. Later, \citet{nazari2018reinforcement} pointed out the limitations of the pointer network and replaced it with simple attention-based embeddings for VRP. \citet{deudon2018learning} and \citet{kool2019attention} proposed to replace the pointer network with the Transformer architecture \cite{vaswani2017attention} called Attention Model (AM) and applied it to solve a variant of problems including TSP and VRP. Recently, \citet{zhang2021} modified the attention model to make it possible to solve the dynamic traveling salesman problem. \citet{xin2021multi} introduced a Multi-Decoder Attention Model (MDAM) to train multiple diverse policies for VRP, which can effectively increase the chance of finding good solutions compared with existing methods which train only one policy.
	
	Heuristic search guided by DRL can be seen as iteration-based. Starting from a complete solution, it is modified to another complete solution by using some search operators at each iteration. The organization and selection of operators are determined by the learning system. \citet{costa2020learning} devised the methods of learning 2-opt heuristics guided by DRL for TSP. \citet{chen2019learning} proposed NeuRewriter that learns a policy to select heuristics and rewrite local components of the current solution to improve it. They tested their approach on job scheduling problem and VRP. \citet{hottung2019neural} proposed a Neural Large Neighborhood Search guided by reinforcement learning for VRPs. Their approach aims to learn how to repair solutions during the LNS procedure. \citet{lu2020a} presented a learning-based iterative method for VRP, which learns to iteratively refine the solution with an improvement operator, selected by a reinforcement learning based controller. \citet{Zhao2020Hybrid} combined a DRL model with a local search method to improve the solution quality. They evaluated their approach on VRPTW instances with different sizes. \citet{fu2021generalize} proposed a novel method that combines machine learning and heuristic techniques such as Monte Carlo tree search, graph sampling, graph converting and heat maps merging, so as to generalize a small pre-trained model to arbitrarily large TSP instances. \citet{zheng2021combining} proposed a variable strategy reinforced approach, denoted as VSR-LKH, which combines three reinforcement learning methods (Q-learning, Sarsa and Monte Carlo) with the well-known Lin-Kernighan-Helsgaun (LKH) algorithm for solving TSP.

	The above discussions have reflected the popularity of recent DRL approaches in solving TSP, VRP and their extensions. It would be even interesting to apply DRL to solve multiobjective versions of these problems. To date, there are several surveys about how to apply RL to handle MOPs (e.g., \cite{peter2011,liu2015}), in which representative multiobjective RL approaches are comprehensively reviewed. However, most of these approaches are unrelated to ``deep" models. To our best knowledge, the existing works only have a few attempts to adopt DRL for MOPs. For example, \citet{li2020deep} and \citet{wu2020modrl} used weight decomposition to transform an MOP into multiple single-objective subproblems, which are solved by DRL with parameter transferring techniques.

	\section{Preliminary}
	\label{2}
	In this section, we briefly discuss some recent DRL methods with weight decomposition for MOPs, particularly MOTSP and MOVRPTW.
	
	\subsection{Benchmark Problems}
	\label{problem}
	
	\subsubsection{MOTSP}
	It is defined on a complete graph with $n$ nodes and $m$ cost matrices \cite{lust2010multiobjective}. The $k$-th cost matrix $[c_{i,j}^k]_{n\times n}$ gives a specific traveling cost from node $i$ to node $j$. The goal is to find a permutation $\pi$ of $n$ nodes to minimize $m$ objective functions simultaneously. The $k$-th objective function is calculated as:
	\begin{equation}
		\small
		\begin{aligned}
			f_k(\pi) = \sum^{n-1}_{i=1}c^k_{\pi(i),\pi(i+1)} + c^k_{\pi(n),\pi(1)}, \quad
			&k=1, 2, \dots,m,
		\end{aligned}
		\label{eq:motsp}
	\end{equation}
	where $\pi(i)$ is the $i$-th element of $\pi$.
	
	The traditional single-objective TSP is a well-known NP-hard problem. It appears that its multiobjective version is even harder. Hence, approximate algorithms are commonly introduced to find near-optimal solutions of MOTSP.
	
	\subsubsection{MOVRPTW}
	
	VRPTW is an extension of TSP by introducing multiple vehicle routes and time window constraints. It shows high relevance with many practical applications.
	
	We consider a multiobjective version of VRPTW. It is defined on a complete graph $G=(V,E)$, where $V=\{0,1,\ldots,n\}$ is the node set. Here, node $0$ is the depot and the remaining nodes are customers. Each node $i$ has attributes including its Euclidean coordinates $c_i \in \mathbb{R}^2$, a time window $tw_i \in \mathbb{R}^2$, and a demand $d_i \in \mathbb{R}$ to be satisfied. Note that the time window $tw_i$ gives the earliest and latest of the possible arrival time at node $i$. For the depot, $q_0=0$ and $tw_0$ essentially corresponds to the earliest departure and latest returning time. $E=\{e_{ij}|i,j\in V, i\neq j\}$ is the edge set. The traveling time on $e_{ij}$, denoted as $t_{ij}$, is set to, for example, the Euclidean distance between $i$ and $j$.
	
	There is a fleet of $K$ uniform vehicles with an identical capacity $Q$. Each vehicle departures from the depot, serves customers under the capacity and time window constraints, then goes back to the depot. The sequence of customers served by vehicle $k$ constitutes a route $r_k$. A solution is a set of routes, denoted as $\pi=\{r_1,r_2,...,r_K\}$. A solution $\pi$ needs to meet the following constraints: (1) all the customers are visited exactly once, i.e., $\bigcup_{i=1}^{K} r_i=V$ and $r_i\bigcap r_j=\emptyset, \ \forall i\neq j$; (2) all the routes satisfy the capacity constraints, i.e., $\sum_{v\in r_k}d_v \leq Q, \ \forall k=1,\ldots,K$; (3) all the customers are served within their specific time windows.
	
	
	Similar to \citet{castro-gutierrez2011nature}, our work focuses on two conflicting objectives: the total traveling time and makespan, described as follows:
	\begin{equation}
		\label{objectives}
		\begin{split}
			&\mathrm{minimize}\quad f(\pi)=(f_1(\pi),f_2(\pi)),\\
			&f_1(\pi)=\sum_{k=1}^{K}c_k,\\
			&f_2(\pi)=\max(c_1,c_2,...,c_K),\\
			&s.t.\  \pi\in \Pi,
		\end{split}
	\end{equation}
	where $c_k$ is the traveling time of $k$-th route in solution $\pi$. $\Pi$ is the valid solution space. Since VRPTW is a classic optimization problem, we omit its formal mathematical formulation in this paper.

	\subsection{Decomposition Strategy}
	\label{decom}

\begin{figure}[htb]
		\centering
		\includegraphics[width=2.4in]{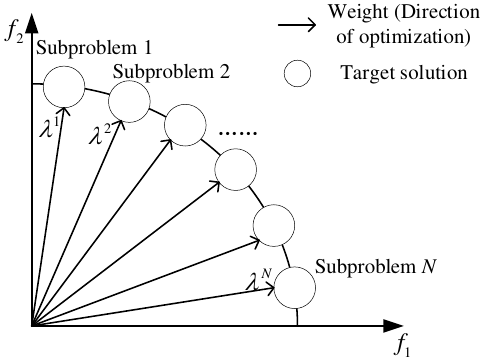}
		\caption{The decomposition strategy of MOP.}
		\label{fig:ws}
	\end{figure}

	Decomposition strategy is a simple yet effective method to design a multiobjective optimization algorithm. As in MOEA/D, MOP is decomposed into $N$ scalar optimization subproblems by
	$N$ weight vectors. Each weight vector corresponds to a particular subproblem \cite{ma2020survey}. The most widely used decomposition approaches include the weighted sum \cite{miettinen2012nonlinear}, Tchebycheff \cite{ma2018Tchebycheff}, and penalty-based boundary intersection (PBI) \cite{zhang2007moea}. In recent DRL methods for MOPs, the decomposition strategy adopts the weighted sum approach which considers the linear combination of different objectives. Specifically, a set of uniformly spread weight vectors $\lambda^{1}, \ldots, \lambda^{N}$ is given, where $\lambda^{j}=(\lambda_{1}^{j}, \ldots, \lambda_{m}^{j})^{T}$ is a weight vector with respect to the $j$-th scalar optimization subproblem, subject to $\lambda_{i}^{j} \geq 0, i=1, \dots, m$ and $\sum_{i=1}^{m} \lambda_{i}^{j} = 1$, e.g., $\lambda^1=(1,0), \lambda^2=(0.9,0.1), \ldots, \lambda^N=(0,1)$ for a bi-objective problem, as shown in \fref{fig:ws}. 
	The objective function of $j$-{th} subproblem, denoted as $f^{ws}(\pi | \lambda^j)$, is given by: 
	\begin{equation}
		\small
		\operatorname{minimize} ~ f^{ws}(\pi | \lambda^j)=\sum_{i=1}^{m} \lambda_{i}^j f_{i}(\pi),\quad j=1,2,\dots,N.
		\label{eq:ws}
	\end{equation}
	
	Solving each scalar optimization problem usually leads to a set of potential Pareto optimal solutions. The desired PF can be obtained when all the $N$ scalar optimization problems are solved.

	\subsection{Framework of DRL for Each Subproblem}
	\label{subframework}
	According to the above decomposition strategy, an MOP is decomposed into a set of subproblems. Each subproblem is treated as a single-objective optimization problem and can be solved by DRL approaches. These approaches are mostly end-to-end and make use of pointer network \cite{pointer}, attention model \cite{nazari2018reinforcement,kool2019attention} or their variants \cite{xin2021multi} as the deep neural networks. The common training algorithms can be policy gradient with rollouts, actor-critic, etc.

	\subsection{Transfer Learning Strategy}
	\label{transfer}
\begin{figure}[htb]
		\centering
		\includegraphics[width=3.5in]{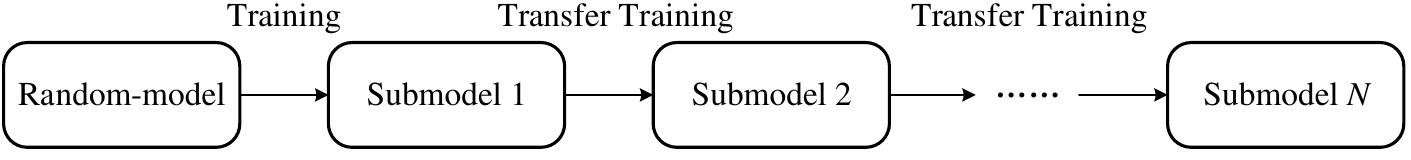}
		\caption{Neighborhood-based transfer learning.}
		\label{fig:ntl}
	\end{figure}

	Since each subproblem requires a lot of training epochs to get its corresponding submodel,  \citet{li2020deep} adopted a transfer learning strategy to speed up the training. It is neighborhood-based, which means that the objective weight vector between two transferring submodels is close to each other. To begin with, the first submodel is trained from scratch with heavy computing power. Then, the neighboring submodel is trained with the initial parameters transferred from the previous submodel. Finally, all the submodels are obtained corresponding to all the weights (see \fref{fig:ntl}).
	
	The shortcoming of the transfer learning strategy is about the flexibility. It is only suitable for training the submodels with their weight vectors closely adjacent to each other. Moreover, all the parameters of submodels need to be saved. It is less applicable when a rich set of new subproblems arises or when the number of objectives is large. In the latter case, any two subproblems could be distant from each other with respect to their objective weight vectors. Hence, such shortcoming motivates us to propose a new learning paradigm for MOPs.
	
	\section{Solution Framework}
	\label{3}
	
	The idea of the proposed MLDRL for solving MOPs is to train a meta-model which can quickly adapt to optimize an unseen objective function corresponding to a new weight. The meta-model, once is trained availably, does not contribute to the construction of PF by itself. Instead, it is used as an optimal initial policy to fine-tune for different submodels with a few more gradient updates. The PF is constructed by aggregating the solutions generated by the submodels.
	
\begin{figure}[t]
		\centering
		\includegraphics[width=3.5in]{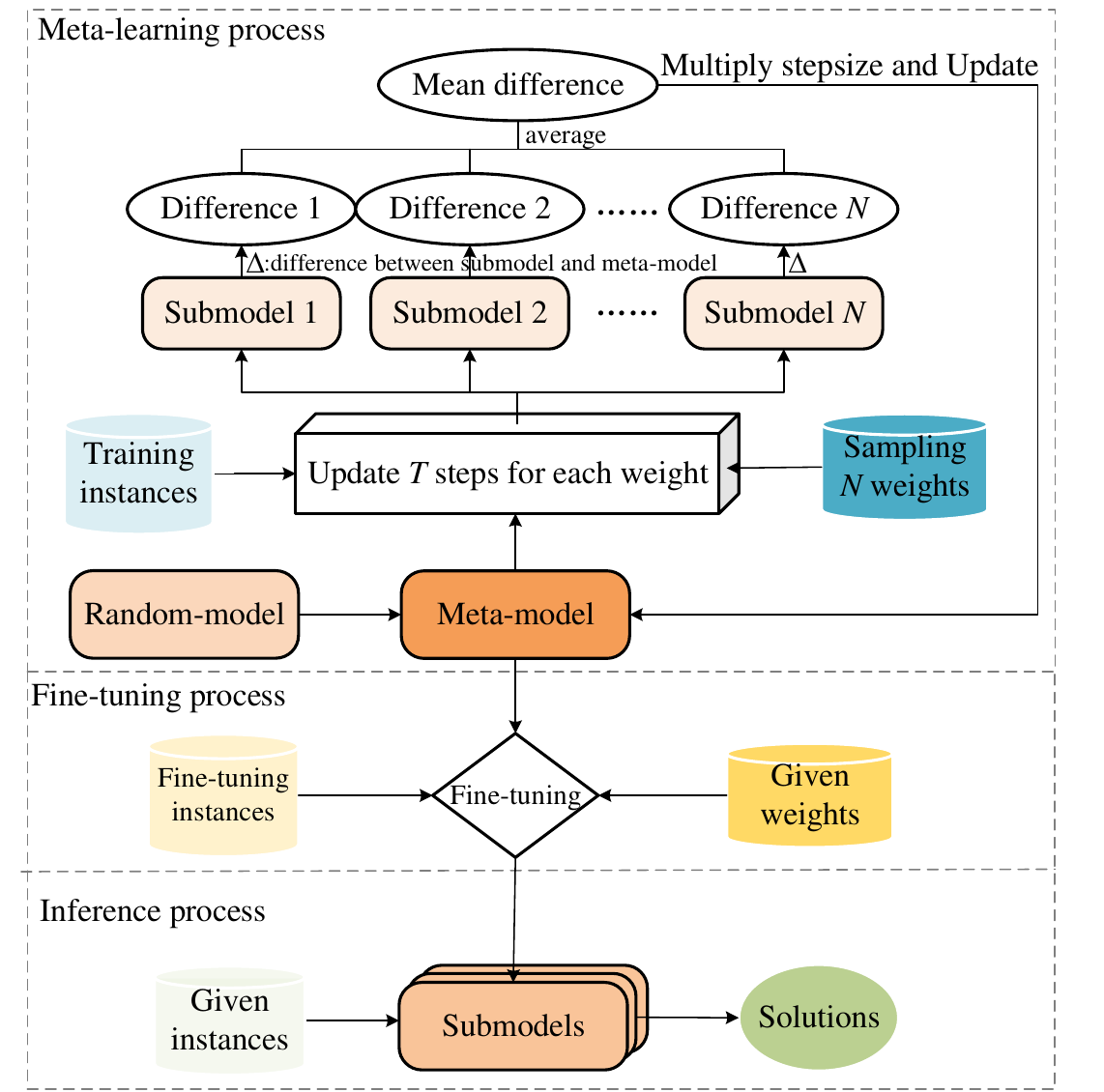}
		\caption{The proposed MLDRL framework.}
		\label{fig:ML}
	\end{figure}

	In short, the proposed MLDRL framework is concise, consisting of a meta-learning process, fine-tuning process and inference process, as illustrated in \fref{fig:ML}. Detailed explanations are presented as follows.
	
	\subsection{Model-agnostic Meta-learning}
	A model-agnostic meta-learning algorithm (MAML) was introduced by \citet{finn2017model}, it is compatible with any model trained with gradient descent and applicable to a variety of learning tasks. The standard MAML requires a second-order derivative, which may consume too many computing resources to train a meta-model. The first-order MAML (FOMAML) would be more efficient, and it has been applied to the continuous motion control field with multiobjective DRL \cite{chen2019meta}. However, FOMAML still needs a training-test split for each learning task.
	
	Recently, a new first-order gradient-based meta-learning algorithm, named Reptile, was proposed by \citet{nichol2018first}. The effect of Reptile is similar to that of FOMAML, but it can save more computing resources due to no back-propagation in the outer loop of training. Furthermore, its training set does not need to be divided into the support set and query set in a specific implementation. Therefore, Reptile would be a more natural choice in our problem settings and is finally chosen as the meta-learning framework for MOPs.

	\subsection{Meta-learning Process}

	The meta-learning process is shown in Algorithm \ref{alg:reptile}. First, a random meta-model $\theta$ is initialized. Then it is trained with a certain iterations ($T_{meta}$). For each iteration, a few weight vectors ($\tilde{N}$) are randomly sampled according to some given distribution $\Lambda$. Each weight vector corresponds to a subproblem that requires DRL to update the parameters of its associated submodel. In other words, a submodel is derived from the meta-model with $T$ update steps guided by a particular weight vector. Next, calculate the difference of the parameters between each submodel and the meta-model, and then these $\tilde{N}$ differences are averaged to get the mean difference. Finally, the mean difference is multiplied by the stepsize $\epsilon$ (the learning rate of meta-learning) to update the parameters of the meta-model.
	
	During meta-learning, the meta-model is updated by multiple subproblems constructed by different weight vectors. Since our meta-learning approach is model-agnostic, it is compatible with any model trained with gradient updates. We hereby adopt the widely used Attention Model (AM) \cite{kool2019attention} for both MOTSP and MOVRPTW. AM is based on the encoder-decoder architecture. It learns a stochastic policy $p(\pi|s)$ for generating a solution $\pi$ given an instance $s$. More detailed architecture of AM is discussed in Section \ref{4}. A family of policy gradient approaches is applicable in training AM. To consistent with some previous approaches \cite{li2020deep,wu2020modrl}, we choose to train it with the actor-critic algorithm.
	
	Algorithm \ref{alg:acml} presents a standard actor-critic training for a subproblem. The actor network $\mu$ is exactly the AM. The critic network $\phi$ shares the encoder of AM but replaces its decoder with two fully-connected layers to output an estimated value. The initial parameters of $\mu$ and $\phi$ are transferred from the meta-model $\theta$. The sample rollout strategy is applied to obtain a solution $\pi_i$ of instance $s_i$. $b_i$ is served as the baseline given by the critic network. We choose ADAM \citep{kingma2015adam} as the training optimizer. The final submodel $\theta_j$ is attained by combining $\mu$ and $\phi$ after $T$ update steps.
	
	\begin{algorithm}[t]
		\small
		\caption{The meta-learning algorithm.}
		\label{alg:reptile}
		\begin{algorithmic}[1]
			\REQUIRE {$\theta$: meta-model, $\Lambda$: the distribution over the weights, $\epsilon_0$: initial outer stepsize, $T_{meta}$: the number of meta-learning iterations, $T$: the number of update steps per submodel, $\tilde{N}$: the number of sampled subproblems, $B$: batchsize per subproblem}
			\ENSURE {The trained meta-model $\theta$}
			\STATE $\epsilon \leftarrow \epsilon_0$
			\FOR{$iteration = 1 : T_{meta}$}
			\FOR{$j = 1 : \tilde{N}$}
			\STATE $\lambda^{j} \leftarrow $ SampleWeight($\Lambda$)
			\STATE$\theta_{j}\leftarrow  ActorCritic(\theta, \lambda^{j}, T, B)$ by Algorithm \ref{alg:acml}
			\ENDFOR
			\STATE $\Delta \theta \leftarrow  \frac{1}{\tilde{N}}\sum_{j = 1}^{\tilde{N}}(\theta_{j}-\theta)$
			\STATE $\theta \leftarrow \theta + \epsilon \Delta  \theta  $
			\STATE $\epsilon \leftarrow \epsilon-\epsilon_0\frac{1}{T_{meta}}$
			\ENDFOR
		\end{algorithmic}
	\end{algorithm}
	
	\begin{algorithm}[t]
		\small
		\caption{Actor-critic algorithm for the $j^{th}$ subproblem.}
		\label{alg:acml}
		\begin{algorithmic}[1]
			\REQUIRE {$(\theta, \lambda^j, T, B)$ from Algorithm \ref{alg:reptile}}
			\ENSURE {The trained submodel $\theta_j$}
			\STATE $\mu, \phi \leftarrow$ DeepCopy($\theta$) ~~~// $\mu$: actor network, $\phi$: critic network
			\STATE Generate a set $\Phi$ of $T\cdot B$ instances 
			
			\FOR{$step = 1 : T$}
			
			\FOR{$i = 1 : B$}
			\STATE $s_i \leftarrow$ SampleInstance($\Phi$)\\
			\STATE $\pi_i \leftarrow$ SampleSolution($p_{\mu}(\cdot|s_i)$) ~// Use SampleRollout\\
			\STATE $b_i \leftarrow$ $b_{\phi}(s_i)$
			
			\ENDFOR
			\STATE$d_\mu \leftarrow \frac{1}{B} \sum_{i=1}^{B}[(f^{ws}(\pi_i|\lambda^j)-b_i)\nabla_{\mu}\log p_{\mu}(\pi_i|s_i)]$ \\
			\STATE$\mathcal{L}_{\phi} \leftarrow \frac{1}{B} \sum_{i = 1}^{B}(b_i-f^{ws}(\pi_i|\lambda^j))^2$ \\
			\STATE$\mu \leftarrow$ ADAM($\mu, d_\mu$) \\
			\STATE$\phi \leftarrow$ ADAM($\phi, \nabla_{\phi}\mathcal{L}_{\phi}$)
			
			\ENDFOR
			\STATE $\theta_j \leftarrow [\mu,\phi]$
		\end{algorithmic}
	\end{algorithm}
	
	\subsection{Fine-tuning and Inference}
	
	The goal of meta-learning is to ensure that every aggregated objective function could be minimized after a small number of gradient updates. Once the meta-model is trained, given any weight vector, we only need to fine-tune the meta-model to obtain a satisfactory submodel that can be used to build or to update the PF.
	
	Algorithm \ref{alg:finetuning} gives how to obtain PF by fine-tuning the meta-model. By using the decomposition strategy discussed in Section \ref{decom}, $N$ weight vectors are generated to define $N$ single-objective problems. For each weight vector, after a few update steps by Algorithm \ref{alg:acml}, the final submodel is obtained. Lines 5--11 correspond to the inference stage. It works for a batch of $K$ test cases. Each case finds its corresponding PF. A solution with respect to the corresponding weight vector can be produced by using the greedy rollout of the policy defined by $\mu_j$. Thereafter, the PF is updated accordingly.
	
	In contrary with previous learning-based methods which directly train a set of submodels, our method trains the meta-model and fine-tunes it to derive submodels. The number of decomposed weight vectors $N$ can be set arbitrarily to get a variety of solutions to constitute the PF. Hence, our method exhibits a better flexibility and versatility.
	
	\begin{algorithm}[t]
		\small
		\caption{Fine-tuning of the meta-model to obtain PF.}
		\label{alg:finetuning}
		\begin{algorithmic}[1]
			\REQUIRE {$\theta$: the well-trained meta-model, $s_i$: the $i$-th instance, $K$: the number of test instances, $N$: the number of decomposed weight vectors, $T$: the number of fine-tuning steps, $B$: batch size per subproblem}
			\ENSURE {Pareto fronts $PF_1,...,PF_K$}
			\FOR{$j=1:N$} 
			\STATE $\lambda^{j} \leftarrow $ GetWeight($\Lambda$)
			\STATE$[\mu_j,\phi_j] \leftarrow  ActorCritic(\theta, \lambda^{j}, T, B)$ by Algorithm \ref{alg:acml}
			\ENDFOR
			\STATE $PF_i=\emptyset,~\forall i=1,\dots,K$
			\FOR{$j=1:N$}
			\FOR{$i=1:K$}
			\STATE $\pi_i \leftarrow$ GetSolution($p_{\mu_j}(\cdot|s_i)$) ~// Use GreedyRollout 
			\STATE $PF_i \leftarrow$ Update($PF_i$, $\pi_i$)
			\ENDFOR	
			\ENDFOR
		\end{algorithmic}
	\end{algorithm}

	\section{Attention Model}
	\label{4}
	
	Attention Model (AM) follows the encoder-decoder architecture. The encoder produces an embedding of all the input information, while the decoder generates the visiting sequence $\pi$ for the corresponding input. In this section, we present the AM for the single-objective TSP and VRPTW, respectively. Note that the multiobjective aspects have already been ruled out according to the weight decomposition. 
	
	
	\subsection{Architecture of AM for TSP}
	
	\subsubsection{Encoder}
	\label{sec:tspe}
	
	Similar to the Transformer architecture \cite{vaswani2017attention}, the encoder consists of a fully connected linear layer and $\mathcal{N}$ attention layers, which transform the low $d_x$-dimensional feature vector space (e.g., $d_x=2$) into a high $d_h$-dimensional embedding vector space (e.g., $d_h=128$) for each node in TSP.
	
	Specifically, for each node $i\in \{1, \dots, n\}$, the fully connected linear layer first maps the node feature vector $x_i$ (i.e., the coordinates of node $i$) to $d_h$-dimensional embedding $h_i^0$:
	\begin{equation}
		h_i^0 = W_0x_i +b_0,
	\end{equation}
	where $W_0$ is a trainable transformation matrix and $b_0$ is a trainable bias vector. Then, $\mathcal{N}$ attention layers further embed the node embeddings $\{h_1^0, \dots, h_n^0\}$. Each attention layer consists of a multi-head attention sublayer (MHA) and a fully connected feed-forward sublayer (FF), and each sublayer uses batch normalization~\cite{ioffe2015batch} (BN) and residual
	connection~\cite{he2016deep} for accelerating the deep network training. The output of each attention layer $\hat{h}_i^l$ can be computed as follows:
	\begin{equation}
		\hat{h}_i^l = \mbox{BN}^{l}(h_i^{l-1} + \mbox{MHA}_i^l(\{h_1^{l-1}, \dots, h_n^{l-1}\}, \mbox{SA})),
	\end{equation}
	\begin{equation}
		h_i^l = \mbox{BN}^l(\hat{h}_i^l + \mbox{FF}^l(\hat{h}_i^l)),
		\label{eq:hil}
	\end{equation}
	where $h_i^l$ represents the embedding of $i$-th node in layer $l$ ($l \in \{1, \dots, \mathcal{N}\}$), MHA is used to aggregate different types of messages from other nodes and its process is shown as follows:
	\begin{equation}
		\mbox{MHA}_i^l(\{h_1^{l-1}, \dots, h_n^{l-1}\}, \mbox{SA}) = \sum_{m = 1}^MW_{m}^{Ol}h_{im}^l,
	\end{equation}
	where $M$ represents the number of heads in MHA, $h_{im}^{l}$ represents the embedding of the $m$-th head of the $i$-th node in the $l$-th attention layer. Note that MHA in the encoder is based on self-attention (SA) and its computing process is described as follows (for convenience, the following four formulas omit $m$ and $l$ for each head in each layer, e.g., $q_{i}$ is short for $q_{im}^{l}$): 
	\begin{equation}
		q_i = W_m^{Ql} h_{im}^{l-1}, \quad k_i = W_m^{Kl} h_{im}^{l-1}, \quad v_i = W_m^{Vl} h_{im}^{l-1},
	\end{equation}
	\begin{equation}
		u_{ij} = \begin{cases}
			\frac{q_i^T k_j}{\sqrt{d_{k}}} & \text{if $i$ adjacent to $j$} \\
			-\infty & \text{otherwise,}
		\end{cases}
	\end{equation}
	\begin{equation}   
		a_{ij} = \frac{e^{u_{ij}}}{\sum_{j'=1}^ne^{u_{ij'}}},
	\end{equation}
	\begin{equation}
		h_{im}^l = \sum_{j=1}^na_{ij}v_{j},
	\end{equation}
	where $d_q = d_k = d_v = \frac{d_h}{M}$, $W_m^{Ql}\in \mathbb{R}^{d_q \times d_h}, W_m^{Kl} \in \mathbb{R}^{d_k \times d_h}, W_m^{Vl} \in \mathbb{R}^{d_v \times d_h}, W_{m}^{Ol} \in \mathbb{R}^{d_h \times d_v}$ are trainable attention matrices of the $m$-th head in the $l$-th attention layer.
	
	FF in Equation (\ref{eq:hil}) consists of a fully-connected layer with ReLu activation function and another fully-connected linear layer: 
	\begin{equation}
		\mbox{FF}^l(\hat{h}_i^l) = W_2^{l}ReLu(W_1^{l}\hat{h}_i^l+b_1^{l}) + b_2^{l},
	\end{equation}
	where $W_1^{l} \in \mathbb{R}^{d_f \times d_h}, W_2^{l} \in \mathbb{R}^{d_h \times d_f}, b_1^l \in \mathbb{R}^{d_f}$ and $b_2^l \in \mathbb{R}^{d_h}$ are trainable parameters. 
	The encoder does not adopt the positional encoding since the input order of nodes is meaningless.
	Finally, we can obtain the node embeddings $\{h_1^\mathcal{N}, \dots, h_n^\mathcal{N}\}$ calculated by $\mathcal{N}$ attention layers and the graph embedding $h_\mathcal{G}^\mathcal{N}$ is defined as follows:
	\begin{equation}
		h_\mathcal{G}^\mathcal{N} = \frac{1}{n}\sum_{i=1}^nh_i^\mathcal{N}.
	\end{equation}
	
	\subsubsection{Decoder}
	\label{sec:dec}
	The decoding process is carried out sequentially. At the $t$-th step of decoding, the decoder outputs the probability $p_\theta(\pi_t = i|\pi_{1:t-1},X)$ of making a decision $\pi_t$ to select a node $i$ of instance $X$ according to the graph embedding $h_\mathcal{G}^\mathcal{N}$ obtained by the encoder and the output information generated at previous time $1,\ldots,t-1$. Given a context embedding $h_{(c)}^\mathcal{N} \in \mathbb{R}^{3d_h}$ calculated by the concatenation of $h_\mathcal{G}^\mathcal{N}$, the first node embedding $h_{\pi_1}^\mathcal{N}$ and the last node embedding $h_{\pi_{t-1}}^\mathcal{N}$, a new context embedding $h_{(c)}^{\mathcal{N}+1}$ is obtained by MHA as follows: 
	\begin{equation}
		h_{(c)}^\mathcal{N} = \begin{cases}[h_\mathcal{G}^\mathcal{N}, h_{\pi_1}^\mathcal{N}, h_{\pi_{t-1}}^\mathcal{N}] \quad t > 1\\ [h_\mathcal{G}^\mathcal{N}, v^b, v^f] \quad \quad \ \  t = 1,\end{cases}
	\end{equation}
	\begin{equation}
		h_{(c)}^{\mathcal{N}+1}=\text{MHA}_{(c)}^{\mathcal{N}+1}(\{h_1^{\mathcal{N}},\cdots,h_n^{\mathcal{N}}\}, h_{(c)}^\mathcal{N}),
		\label{eq:mha}
	\end{equation}
	where $v^b \in \mathbb{R}^{d_h}, v^f \in \mathbb{R}^{d_h}$ are two trainable parameter vectors (placeholders).
	
	Based on the new context embedding $h_{(c)}$, the probability $p_\theta(\pi_t = i|\pi_{1:t-1}, X)$ is calculated by a single-head attention layer: 
	\begin{equation}
		q = W_qh_{(c)}^{\mathcal{N}+1}, \quad k_i = W_kh_i^\mathcal{N},
	\end{equation}
	\begin{equation}
		u_{i} = \begin{cases}C\cdot \mbox{tanh}(q^{T}k_i) \quad \quad \quad i \notin \pi_{1:t-1}\\ -\infty \quad \quad \quad \quad \quad \quad \quad \ \   \mbox{otherwise,}\end{cases}
	\end{equation}
	\begin{equation}
		p_\theta(\pi_t = i|\pi_{1:t-1}, X) = \frac{e^{u_{i}}}{\sum_{j=1}^{n}e^{u_{j}}},
	\end{equation}
	where $W_q \in \mathbb{R}^{d_h \times d_h}$ and $W_k \in \mathbb{R}^{d_h \times d_h}$ are trainable parameters, we set $C=10$ to clip the result~\cite{bello2017neural}. Finally, a solution $\pi$ of instance $X$ can be obtained by $p_{\theta}(\pi|X)$ based on the probability chain rule: 
	\begin{equation}
		p_{\theta}(\pi|X) = \prod ^{n}_{t=1}p_\theta(\pi_t = i|\pi_{1:t-1}, X).
	\end{equation}
	
	\subsection{AM for VRPTW}
	\label{sec:vrp}
	
	\subsubsection{Encoder}
	
	The input of a VRPTW instance contains more information than that of a TSP instance. For each node $i\in\{1,\ldots,n\}$, the feature vector $x_i = (c_i, tw_i, d_i) \in \mathbb{R}^5$, which corresponds to the coordinates, time window and demand of the node. Hence, the dimension $d_x=5$. Other components of the encoder are the same as those in Section \ref{sec:tspe}.

	\subsubsection{Decoder}
	In the decoder of VRPTW, the context embedding $h_{(c)}^\mathcal{N} \in \mathbb{R}^{2d_h+2}$ is calculated by the concatenation of $h_\mathcal{G}^\mathcal{N}$, the last served node embedding $h_{\pi_{t-1}}^\mathcal{N}$, the remaining capacity $D_{t-1}$, and the current time $T_{t-1}$ after serving the last node. Then, a new context embedding $h_{(c)}^{\mathcal{N}+1}$ is obtained by MHA, as follows:

	\begin{equation}
		h_{(c)}^\mathcal{N} = [h_\mathcal{G}^\mathcal{N},  h_{\pi_{t-1}}^\mathcal{N}, D_{t-1}, T_{t-1}], \quad t \geq 1,
	\end{equation}
	\begin{equation}
		h_{(c)}^{\mathcal{N}+1}=\text{MHA}_{(c)}^{\mathcal{N}+1}(\{h_1^{\mathcal{N}},\cdots,h_n^{\mathcal{N}}\}, h_{(c)}^\mathcal{N}).
	\end{equation}
	
	Here, $D_0$ and $T_0$ are the initial total capacity and the initial time of the vehicle respectively. $D_t$ and $T_t$ are dynamically updated at each decoding step. Note that if the depot $0$ is selected at some step $t$, a new vehicle is dispatched and we have $D_t = D_0$ and $T_t=T_0$.
	
	The remaining calculation procedure of the decoder is very similar to the one in Section \ref{sec:dec}, from Equation (17) to (20), except that the compatibility variable $u_i$ is set as follows:
	\begin{equation}
		u_{i} = \begin{cases}C\cdot \mbox{tanh}(q^{T}k_i) \quad \quad \quad \mbox{node $i$ is not masked}\\ -\infty ~ \quad \quad \quad \quad \quad \quad \ \   \mbox{otherwise.}\end{cases}
	\end{equation}
	
	The mask mechanism is applied to restrict the value of $u_i$, i.e., if node $i$ is masked, $u_i=-\infty$. Because VRPTW is more complex than TSP, the following constraints should be checked to validate the masking of node $i$.
	\begin{itemize}
		\item Node $i$ has been visited before, i.e., $i\in \pi_{1:t-1}$.
		\item The capacity of the vehicle is exceeded, i.e., $d_i>D_{t-1}$.
		\item The time window constraint of node $i$ is violated, i.e., $T_{t-1}+t_{\pi_{t-1},i}>\overline{tw_i}$, where $\overline{tw_i}$ indicates the right-end of node $i$'s time window.
	\end{itemize}

	\section{Experiments: MOTSP}
	\label{5}
	
	\subsection{Experiment Settings}
	
	All the experiments were conducted on a machine with an Intel Xeon E5-2637 CPU and several TITAN-Xp GPUs. The compared algorithms, including three learning-based approaches and three iteration-based approaches, are summarized in Table \ref{tab:alg1}. Our codes are available at \url{https://github.com/zhangzizhen/ML-DAM}.

	\subsubsection{Test Instances}
	The proposed method was tested on a classic MOTSP called Euclidean bi-objective TSP \cite{lust2010multiobjective}. This problem includes two objectives that need to be optimized simultaneously, each of which considers an independent Euclidean distance matrix. A node in the problem has two coordinate features, both uniformly sampled from $[0, 1]\times [0, 1]$ or normalized to $[0, 1]\times [0, 1]$ from standard benchmark instances. Each coordinate feature is used to calculate the Euclidean distance for its corresponding objective.
	

	\begin{table}[!h]
		\centering
		\small
		\caption{Detailed descriptions of the compared algorithms for MOTSP.}
		\begin{tabular}{p{0.15\columnwidth}p{0.75\columnwidth}}
			\toprule
			Algorithm & Description \\
			\midrule
			ML-AM & Meta-Learning-based AM for MOTSP. \\
			AM & AM-based DRL for each of $N=100$ TSP subproblems. \\
			AM-T & AM with transfer-learning adopted in \citet{li2020deep}. \\
			NSGA-II & Standard Non-Dominated Sorting Genetic Algorithm II on the platform PlatEMO\footnotemark[1] \cite{PlatEMO}. \\
			MOEA/D & Standard Multi-Objective Evolutionary Algorithm with Decomposition on the platform PlatEMO. \\
			MOGLS & Multi-Objective Genetic Local Search implemented with Python\footnotemark[2] according to \citet{jaszkiewicz2002genetic}. \\
			\bottomrule
		\end{tabular}%
		\label{tab:alg1}%
	\end{table}%
	\footnotetext[1]{\url{https://github.com/BIMK/PlatEMO}}
	\footnotetext[2]{\url{https://github.com/kevin031060/Genetic\_Local\_Search\_TSP}}

	\subsubsection{Subproblems for Constructing PF}
	We set the number of weight vectors $N$ in PF construction to $100$. They are uniformly spread from $(0,1)$ to $(1,0)$, i.e., $\lambda^{1}=(0, 1)$, $\lambda^{2}=(\frac{1}{99}, \frac{98}{99})$, ..., $\lambda^{100}=(1, 0)$. 
	
	\subsubsection{Hyperparameters of AM}
	
	In order to make a fair comparison of our method with previous learning-based methods, the network model of every subproblem uses the same AM architecture as proposed in \citet{kool2019attention}. The input dimension $d_x$ is set to 4 (i.e., two coordinates) for MOTSP and the dimension of node embedding $d_h$ is set to 128. The number of multi-head attention (MHA) layers for the encoder is set to 3. For each MHA layer, the number of heads $A$ is set to 8 and the dimensions $d_q$, $d_k$ and $d_v$ are all set to $\frac{d_h}{A}$ = 16.

	\subsubsection{Training Details}
	We use the Adam optimizer for the parameter update of each submodel in the inner loop. The learning rate is set to ${10}^{-4}$ constantly. After some preliminary experiments, the number of update steps $T$ per submodel is set to $100$ and the number of submodels $\tilde{N}$ per meta-model update is set to $5$. The batch size $B$ of instances is set to $512$ per subproblem. The meta-learning rate $\epsilon$ is linearly annealed to zero during meta-training with the initial value $\epsilon_0=1.0$. 
	
	We have conducted the training of meta-models on randomly generated MOTSP instances with 20 nodes (MOTSP-20) and 50 nodes (MOTSP-50), respectively. The iteration of meta-learning $T_{meta}$ is set to 10000 and 5000 for MOTSP-20 and MOTSP-50, respectively. 
	
	\subsubsection{Evaluation Metrics}
	
	We use Hypervolume (HV) and the number of non-dominated solutions ($|$NDS$|$) to evaluate the performance of the compared algorithms. HV is an important indicator to comprehensively evaluate the convergence and diversity of PF, while NDS reflects the diversity of PF when HV values are close to each other. In general, a larger HV or NDS indicate a better performance of the corresponding algorithm.

	\subsection{Experimental Results}
	
	\subsubsection{Model Performance in Meta-Learning}
	
	\begin{figure*}[t]
		\centering
		\subfloat[MOTSP-20]{\includegraphics[width=3.5in]{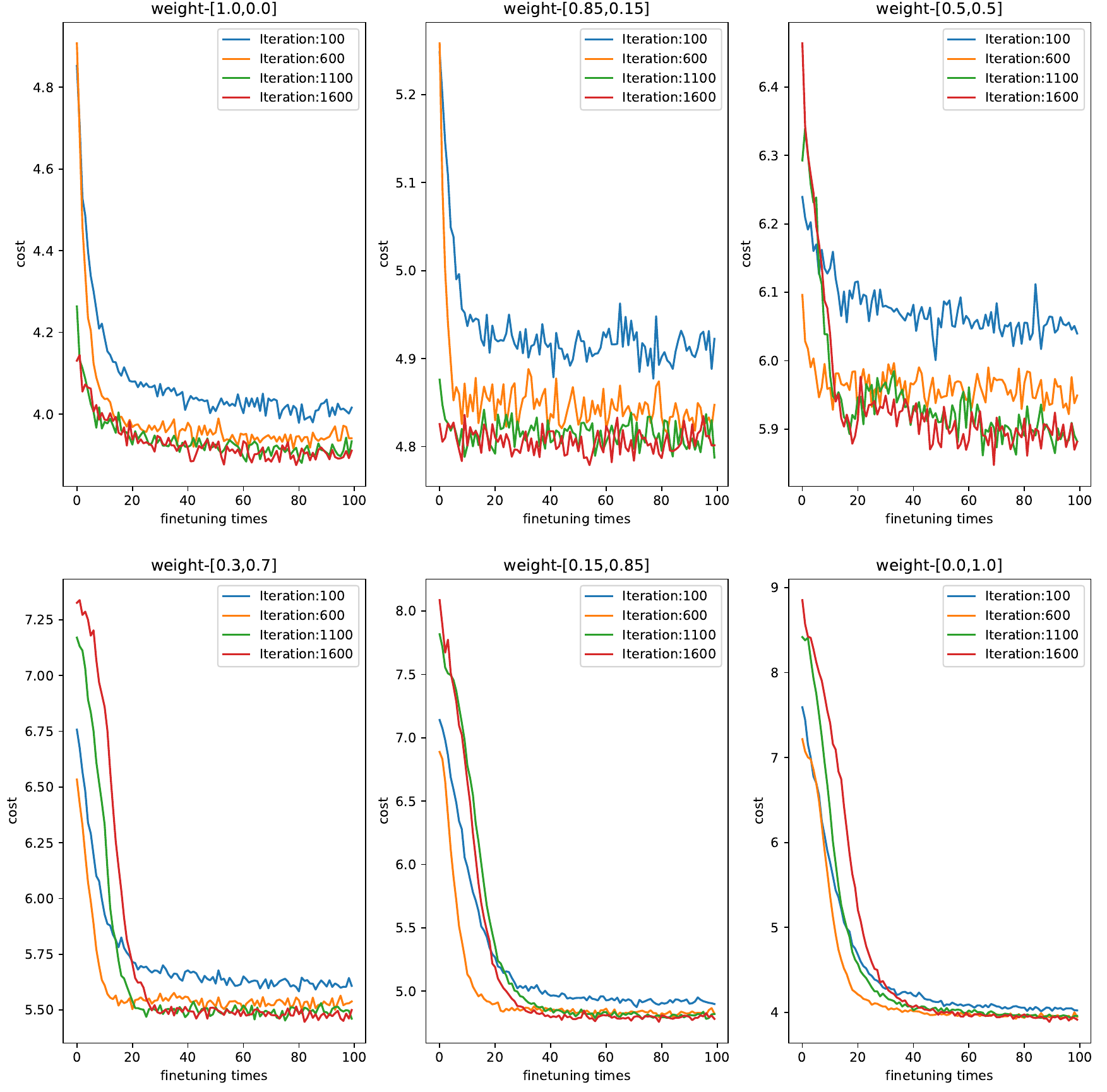}%
		}
		\hfil
		\subfloat[MOTSP-50]{\includegraphics[width=3.5in]{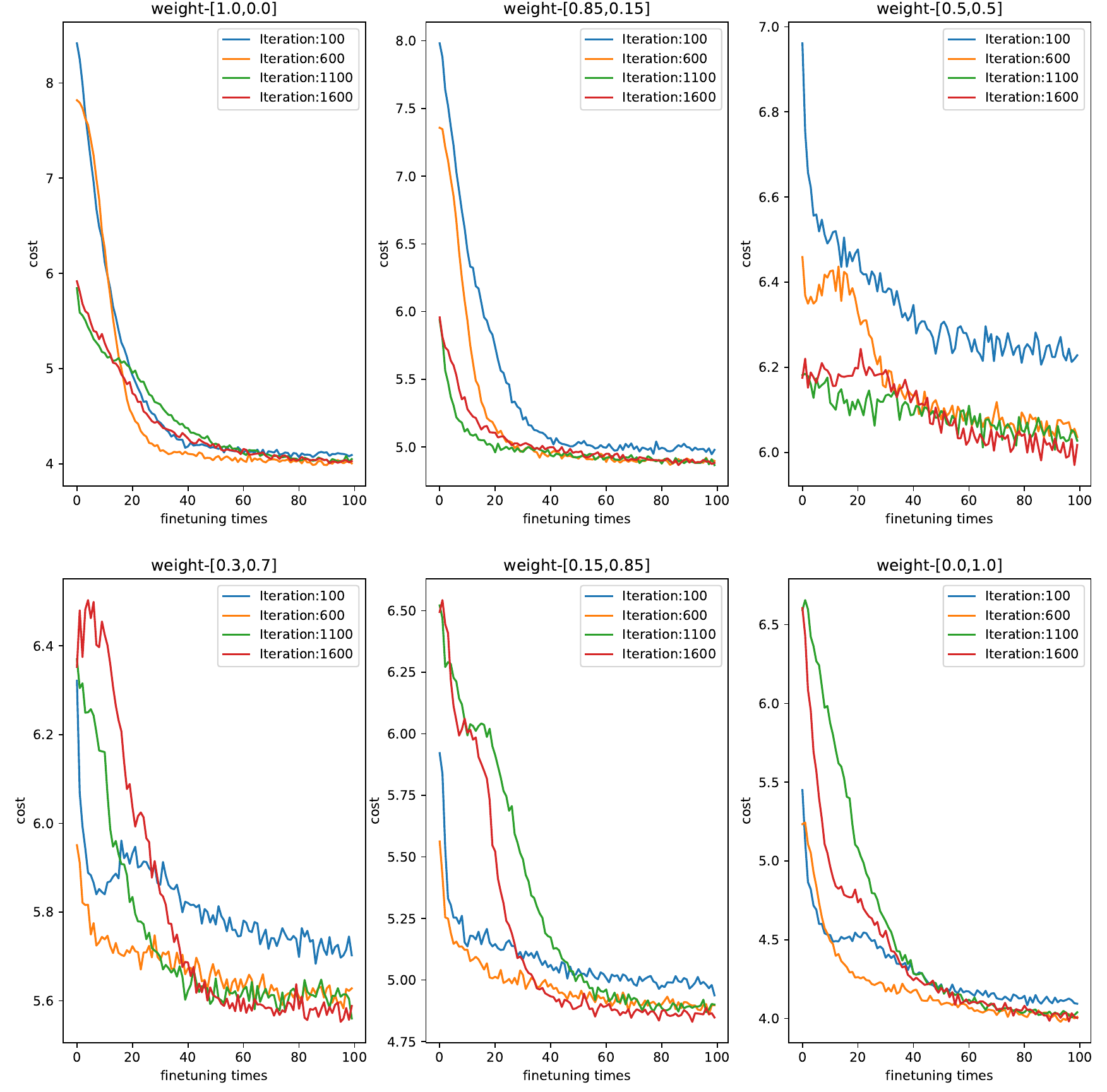}%
		}
		\caption{The performance of the meta-models trained with different iterations.}
		\label{fig:tsp-fintuing}
	\end{figure*}
	
	To evaluate the performance of the meta-model partly trained during the meta-learning process, we saved the models trained with 100, 600, 1100 and 1600 iterations. Then, we used these partly trained models to fine-tune 100 steps under different weight vectors to obtain the corresponding submodels.
	
	\fref{fig:tsp-fintuing} shows the average weighted-sum cost along with the fine-tuning times for MOTSP-20 and MOTSP-50, with respect to different weight vectors and meta-learning iterations. When the fine-tuning steps increase, all the models obtained by different meta-learning iterations can get improved, as all the curves show a downward trend. What is more, a significant drop can be observed in the first 20 fine-tuning steps, which indicates that slightly fine-tuning the meta-model suffices to produce a promising submodel. Compared with the model trained with a few iterations (e.g., 100), the model trained with a plenty of iterations (e.g., 1600) exhibits better convergence when the fine-tuning steps become large. These facts verify the effectiveness of the meta-learning algorithm.

	\subsubsection{Performance of Learning-based Approaches for MOTSP}
	To evaluate the performance of different learning-based approaches, we tested them which were trained with 20 nodes and 50 nodes on 128 random instances of MOTSP-20 and MOTSP-50, respectively. AM and ML-AM with $T$-step mean that we used AM and ML-AM to update the random model and meta-model for each subproblem with $T$ steps, respectively. For AM-T with $T$-step, we first applied AM to update the initial model with $5000$ steps for the first subproblem with weight vector $\lambda^{1}=(0, 1)$. Then the next model is obtained by the previous submodel with $T$ update steps for each of the remaining $99$ subproblems.
	
	Table \ref{tb:20-50-HV} shows the average performance in terms of HV and $|$NDS$|$ on $128$ random instances. The reference point for calculating HV is set to $(30, 30)$ for both MOTSP-20 and MOTSP-50. From this table, we find that by only taking 10 update steps, ML-AM can generate very promising solutions. It significantly outperforms AM with thousands of update steps. Compared with AM-T, ML-AM still shows better performance in terms of both HV and $|$NDS$|$ when the same number of update steps is considered. The results also suggest that ML-AM with a proper update steps (e.g., 100 steps) is competent to produce PF with good convergence and diversity. Since the same network architecture is adopted, the running time of these methods is similar under the same update steps.

\begin{table}[t]
	\centering
	\caption{The average results of HV and $|$NDS$|$ tested on $128$ random instances with 20 and 50 nodes. The higher value is marked in bold.}
	\resizebox{1\columnwidth}{!}{
		\begin{threeparttable}
			\begin{tabular}{|c|c|c|c|c|c|c|c|}
				\hline
				\multirow{2}{*}{Scale} & \multicolumn{1}{c|}{\multirow{2}{*}{Parameter update}} & \multicolumn{2}{c|}{ML-AM} & \multicolumn{2}{c|}{AM} & \multicolumn{2}{c|}{AM-T} \\ \cline{3-8} 
				& \multicolumn{1}{c|}{}                                  & HV               & $|$NDS$|$           & HV          & $|$NDS$|$      & HV                & $|$NDS$|$           \\ \hline
				\multirow{4}{*}{20}    & 10-step                                                & \textbf{668.56}  & \textbf{69.18}  & 223.19      & 7.92       & 653.72            & 25.11           \\ \cline{2-8} 
				& 100-step                                               & \textbf{672.19}  & \textbf{60.91}  & 332.28      & 14.46      & 670.21            & 39.68           \\ \cline{2-8} 
				& 1000-step                                              & \textbf{672.50}  & \textbf{56.48}  & 448.22      & 26.66      & 671.91            & 46.91           \\ \cline{2-8} 
				& 5000-step                                              & \textbf{672.64}  & \textbf{56.62}  & 480.05      & 34.88      & 672.56            & 54.20           \\ \hline
				\multirow{3}{*}{50}    & 10-step                                                & \textbf{496.83}  & \textbf{42.25}  & 219.70      & 7.05       & 419.73            & 24.04           \\ \cline{2-8} 
				& 100-step                                               & \textbf{504.74}  & \textbf{48.93}  & 328.25      & 13.45      & 484.47            & 35.69           \\ \cline{2-8} 
				& 1000-step                                              & \textbf{504.75}  & \textbf{47.70}  & 448.38      & 26.80      & 500.28            & 42.86           \\ \hline
			\end{tabular}
			\begin{tablenotes}
				\footnotesize
				\item[] Note: For each step of the parameter update of a submodel, any of the three algorithms takes about 0.3 seconds for TSP-20 and 1.1 seconds for TSP-50.
			\end{tablenotes}
		\end{threeparttable}
	}
	
	\label{tb:20-50-HV}
\end{table}

\subsubsection{Generalization Ability of the Meta-model}
In order to test the generalization ability of the meta-model to adapt to different learning tasks, we applied the meta-model trained with MOTSP-50 and fine-tuned on random instances of MOTSP-30, MOTSP-80 and MOTSP-100, respectively. In comparison, AM and AM-T were directly trained with MOTSP-30, MOTSP-80 and MOTSP-100, respectively. The average results of HV and $|$NDS$|$ on $128$ random instances are shown in Table~\ref{tb:30-80-100-HV}. The reference point for calculating HV is set to $(60, 60)$ for these instances. 

The reported results indicate a powerful generalization ability of the meta-model in adapting to other problems with different scales. Although AM and AM-T were directly trained on those related problem instances without incurring the scalability issue, they are still inferior to ML-AM. The fact suggests that the meta-model indeed captures some common features of MOTSP, slightly fine-tuning the meta-model suffices to produce satisfactory results for different scale instances.


\begin{table}[t]
	\centering
	\caption{The average results of HV and $|$NDS$|$ to evaluate the generalization ability of the meta-model. The test suite has $128$ random instances with different scales. The higher value is marked in bold.}
	\resizebox{1\columnwidth}{!}{
		\begin{tabular}{|c|c|c|c|c|c|c|c|}
			\hline
			\multirow{2}{*}{Scale} & \multirow{2}{*}{Parameter update} & \multicolumn{2}{c|}{ML-AM} & \multicolumn{2}{c|}{AM} & \multicolumn{2}{c|}{AM-T} \\ \cline{3-8} 
			&                                   & HV                & $|$NDS$|$          & HV           & $|$NDS$|$     & HV                & $|$NDS$|$           \\ \hline
			\multirow{3}{*}{30}    & 10-step                           & \textbf{3003.69}  & \textbf{42.50} & 2591.97      & 7.04      & 2999.11           & 22.66           \\ \cline{2-8} 
			& 100-step                          & \textbf{3037.67}  & \textbf{49.36} & 2853.46      & 12.54     & 3032.39           & 33.26           \\ \cline{2-8} 
			& 1000-step                         & \textbf{3040.90}  & \textbf{48.39} & 2995.74      & 22.98     & 3039.51           & 41.63           \\ \hline
			\multirow{3}{*}{80}    & 10-step                           & \textbf{2473.44}  & \textbf{43.42} & 1362.92      & 6.44      & 2284.43           & 26.99           \\ \cline{2-8} 
			& 100-step                          & \textbf{2560.56}  & \textbf{52.30} & 1861.99      & 12.91     & 2496.17           & 40.23           \\ \cline{2-8} 
			& 1000-step                         & \textbf{2569.88}  & \textbf{55.06} & 2315.11      & 29.20     & 2548.81           & 47.80           \\ \hline
			\multirow{3}{*}{100}   & 10-step                           & \textbf{2258.21}  & \textbf{43.30} & 978.91       & 8.28      & 1935.39           & 28.48           \\ \cline{2-8} 
			& 100-step                          & \textbf{2343.70}  & \textbf{53.25} & 1444.58      & 12.80     & 2256.31           & 42.27           \\ \cline{2-8} 
			& 1000-step                         & \textbf{2370.19}  & \textbf{58.18} & 2039.99      & 30.89     & 2330.12           & 49.63           \\ \hline
		\end{tabular}
	}
	\label{tb:30-80-100-HV}
\end{table}

\subsubsection{Versatility of the Meta-model}

\begin{figure}[t]
	\centering
	\includegraphics[width=3.5in]{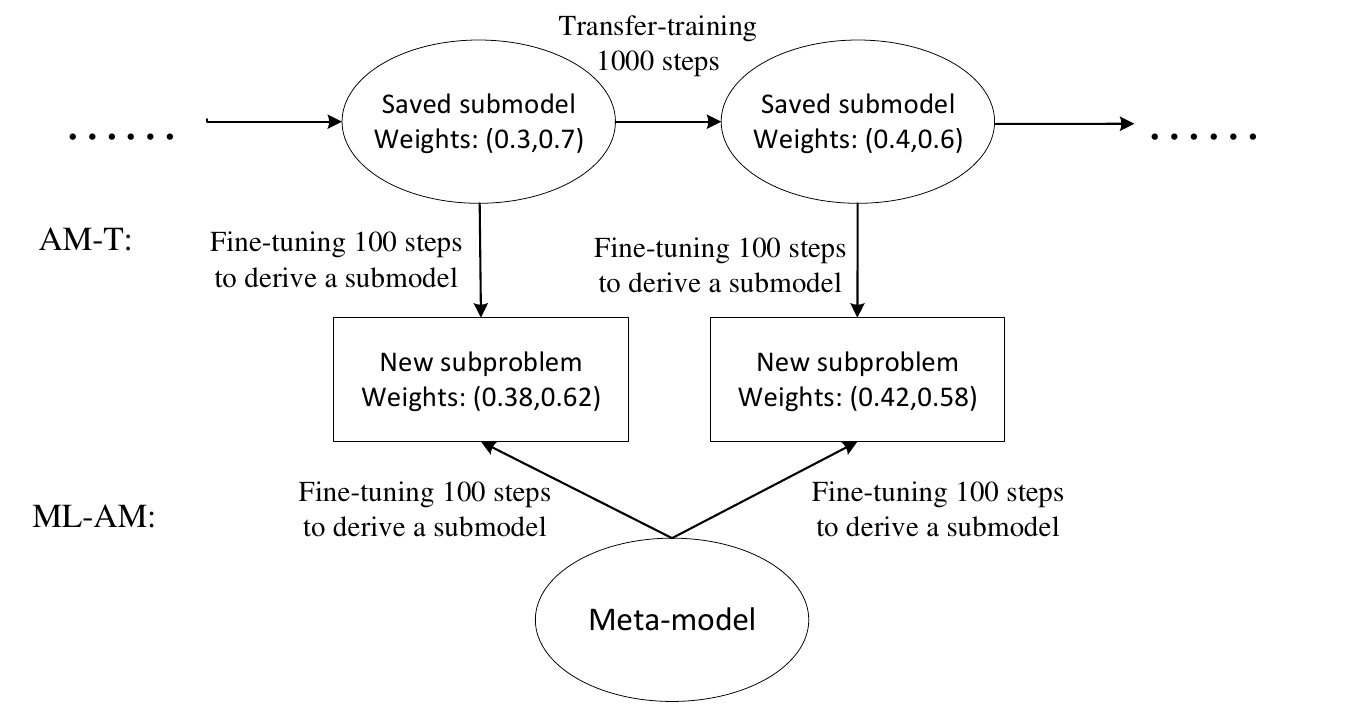}
	\caption{The illustration of ML-AM and AM-T in handling new subproblems with unseen weight vectors.}
	\label{fig:unseen}
\end{figure}

\begin{table}[t]
	\centering
	\caption{The comparison results of ML-AM and AM-T in handling new subproblems with unseen weight vectors. The smaller $f^{ws}$ is marked in bold.}
	\resizebox{1\columnwidth}{!}{
		\begin{tabular}{|c|c|c|c|c|c|c|c|}
			\hline
			\multirow{1}[0]{*}{Weight vector} & \multicolumn{3}{c|}{ML-AM} & \multicolumn{4}{c|}{AM-T} \\
			\cline{2-8}
			$(\lambda_1,\lambda_2)$ & $f_1$  & $f_2$  & $f^{ws}$     & Submodel & $f_1$  & $f_2$  & $f^{ws}$ \\
			\hline
			(0.02,0.98) & 35.69 & 8.22  & \textbf{8.77} & (0.0,1.0) & 41.65 & 9.42  & 10.06 \\ \hline
			(0.09,0.91) & 34.62 & 8.10  & \textbf{10.49} & (0.0,1.0) & 43.24 & 9.98  & 12.97 \\ \hline
			(0.11,0.89) & 32.75 & 9.04  & \textbf{11.64} & (0.1,0.9) & 44.00 & 8.01  & 11.97 \\ \hline
			(0.18,0.82) & 32.28 & 9.38  & \textbf{13.50} & (0.1,0.9) & 42.68 & 8.12  & 14.34 \\ \hline
			(0.28,0.72) & 23.51 & 11.19 & \textbf{14.64} & (0.2,0.8) & 44.33 & 8.88  & 18.81 \\ \hline
			(0.29,0.71) & 27.88 & 9.41  & \textbf{14.76} & (0.2,0.8) & 42.94 & 8.36  & 18.39 \\ \hline
			(0.31,0.69) & 22.20 & 10.86 & \textbf{14.38} & (0.3,0.7) & 42.46 & 9.21  & 19.52 \\ \hline
			(0.38,0.62) & 22.01 & 11.91 & \textbf{15.75} & (0.3,0.7) & 44.30 & 9.56  & 22.76 \\ \hline
			(0.42,0.58) & 18.25 & 13.40 & \textbf{15.44} & (0.4,0.6) & 26.24 & 14.15 & 19.23 \\ \hline
			(0.47,0.53) & 17.17 & 14.13 & \textbf{15.56} & (0.4,0.6) & 23.74 & 12.81 & 17.95 \\ \hline
			(0.53,0.47) & 15.75 & 16.86 & \textbf{16.27} & (0.5,0.5) & 19.00 & 18.30 & 18.67 \\ \hline
			(0.56,0.44) & 16.37 & 16.60 & \textbf{16.47} & (0.5,0.5) & 16.90 & 20.65 & 18.55 \\ \hline
			(0.64,0.36) & 14.77 & 17.57 & \textbf{15.78} & (0.6,0.4) & 15.42 & 21.91 & 17.76 \\ \hline
			(0.65,0.35) & 12.03 & 18.23 & \textbf{14.20} & (0.6,0.4) & 14.80 & 22.38 & 17.45 \\ \hline
			(0.74,0.26) & 10.40 & 23.52 & \textbf{13.81} & (0.7,0.3) & 11.69 & 27.73 & 15.86 \\ \hline
			(0.75,0.25) & 10.05 & 22.99 & \textbf{13.29} & (0.7,0.3) & 11.70 & 26.43 & 15.38 \\ \hline
			(0.83,0.17) & 8.16  & 30.68 & \textbf{11.99} & (0.8,0.2) & 9.83  & 35.89 & 14.26 \\ \hline
			(0.86,0.14) & 8.10  & 30.37 & \textbf{11.22} & (0.8,0.2) & 10.19 & 38.83 & 14.20 \\ \hline
			(0.92,0.08) & 8.59  & 31.61 & \textbf{10.43} & (0.9,0.1) & 8.76  & 44.43 & 11.61 \\ \hline
			(0.97,0.03) & 8.04  & 32.76 & \textbf{8.79} & (0.9,0.1) & 9.33  & 41.04 & 10.28 \\ \hline
		\end{tabular}%
	}
	\label{tab:unseen}%
\end{table}%

The versatility of a model indicates whether it can be reused multiple times. To verify the versatility of the meta-model in handling new subproblems with unseen weight vectors, we conducted some experiments on instance kroAB100. AM-T is selected as the compared algorithm, in which 10 trained submodels (with weight vectors $\lambda^1=(0,1),...,\lambda^{10}=(0.9,0.1)$ via 1000 steps transfer-training) are saved in advance. When a new subproblem with previously unseen weight vector is tested, we use: 1) the meta-model to fine-tune 100 steps to derive a submodel to solve the subproblem; 2) the neighboring submodel saved in AM-T to fine-tune 100 steps to derive another submodel to solve the subproblem. \fref{fig:unseen} illustrates the basic processes.

Table \ref{tab:unseen} presents the comparison results. Twenty weight vectors are generated for the testing. The column ``$f^{ws}$" is obtained by $\lambda_1 f_1+\lambda_2 f_2$, which gives the weighted-sum cost for the corresponding subproblem. The column ``Submodel" indicates the neighboring submodel in AM-T for the fine-tuning. As shown, ML-AM outperforms AM-T for all the new subproblems. The reasons are as follows. The submodels trained in AM-T are targeted at specific subproblems. Fine-tuning the neighboring submodels with a few steps may not lead to competent submodels for new subproblems. In comparison, the meta-model in ML-AM has learned from various subproblems and captured their common features. Therefore, it can better handle new subproblems with unseen weight vectors. It is worth noting that ML-AM only stores a single meta-model in memory, while AM-T needs to save ten submodels in this scenario. All the facts show that MLDRL is a good learning paradigm for MOPs.

\subsubsection{Generalization Ability of the Submodel}

To test the generalization ability of the submodels for tackling different scale instances, we applied the submodels derived from the meta-model trained and fine-tuned $T$ steps with MOTSP-50 to three commonly used benchmark MOTSP instances: kroAB100, kroAB150 and kroAB200 \cite{li2020deep}. These instances were constructed by kroA and kroB in the TSP library \cite{TSPLIB}. The reference point for calculating HV is set to $(90, 90)$ for these instances.

\begin{figure*}[!h]
	\centering
	\subfloat[]{\includegraphics[width=2.1in]{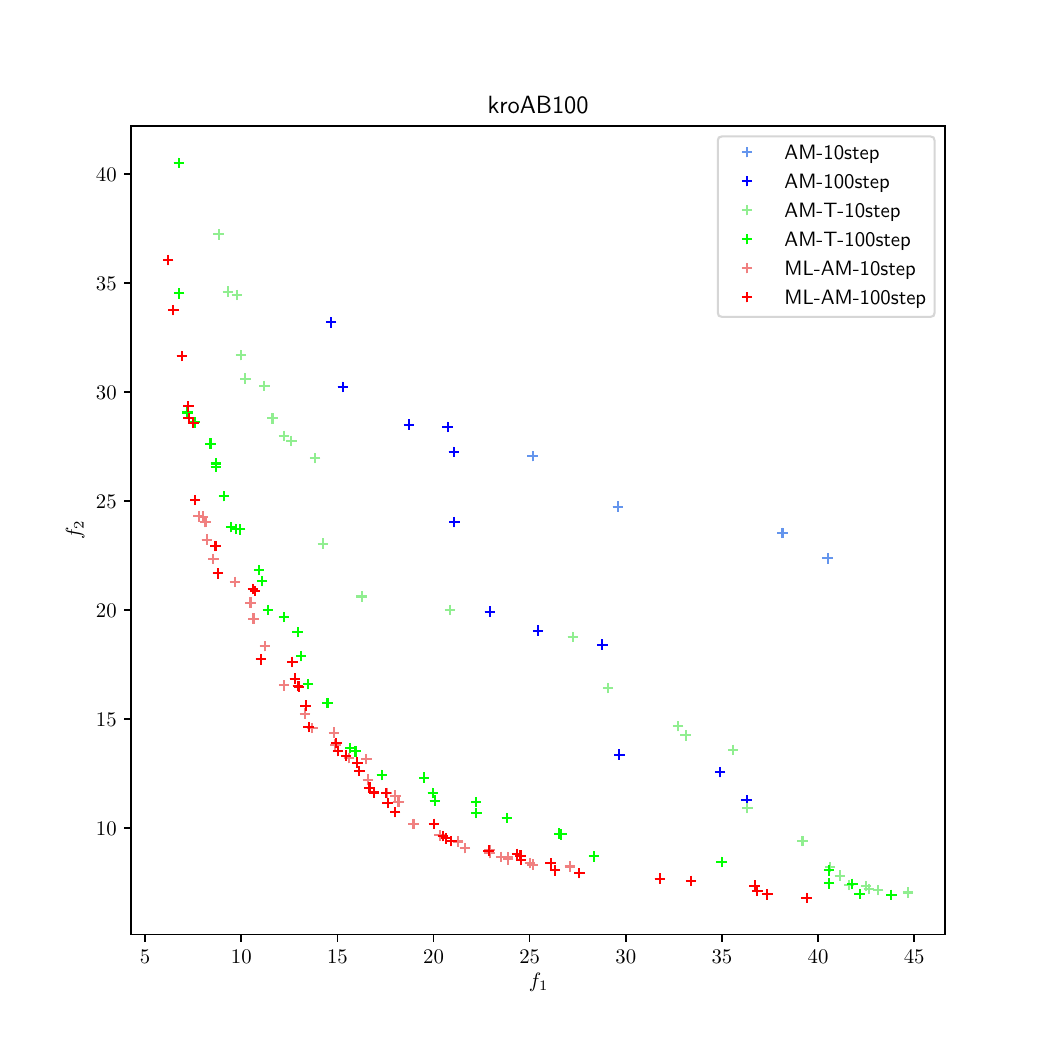}%
	}
	\hfil
	\subfloat[]{\includegraphics[width=2.1in]{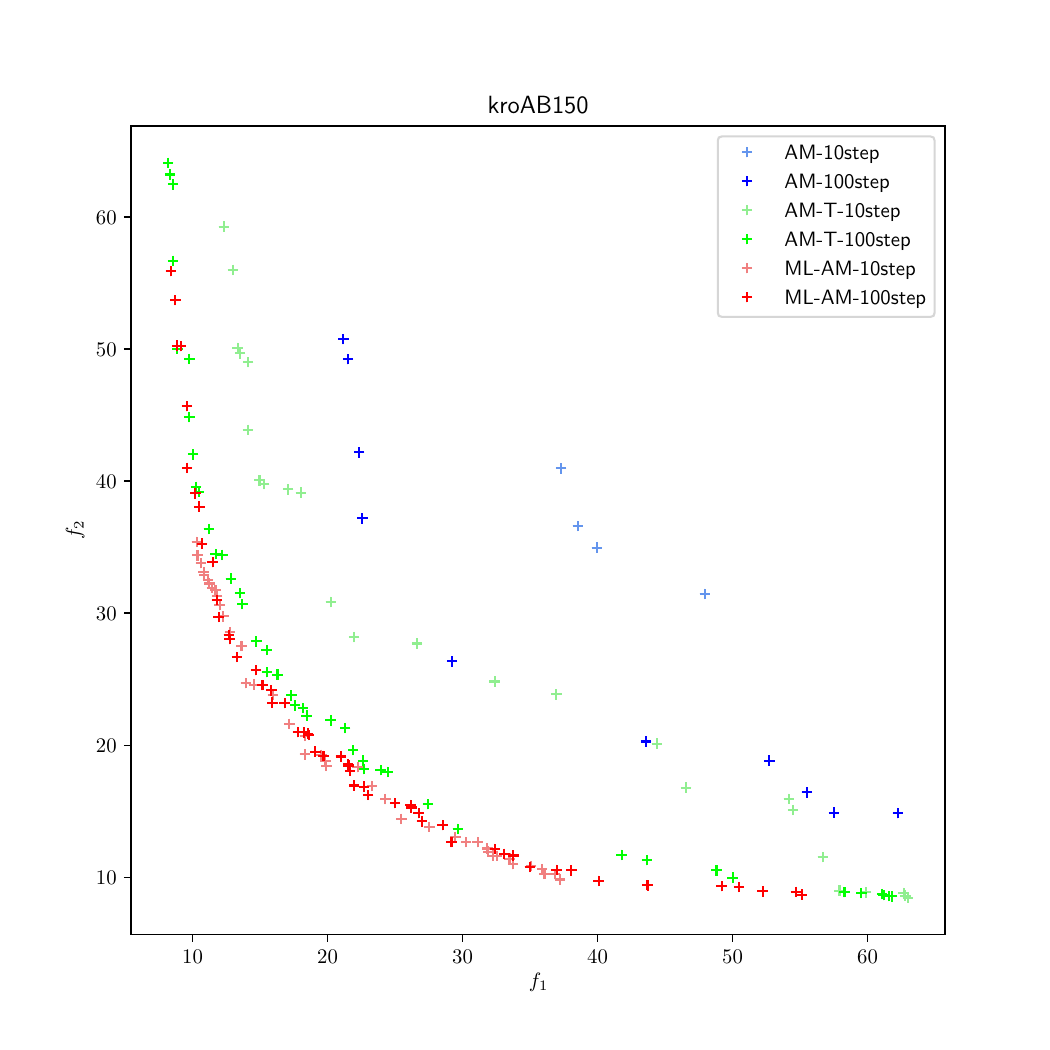}%
	}
	\hfil
	\subfloat[]{\includegraphics[width=2.1in]{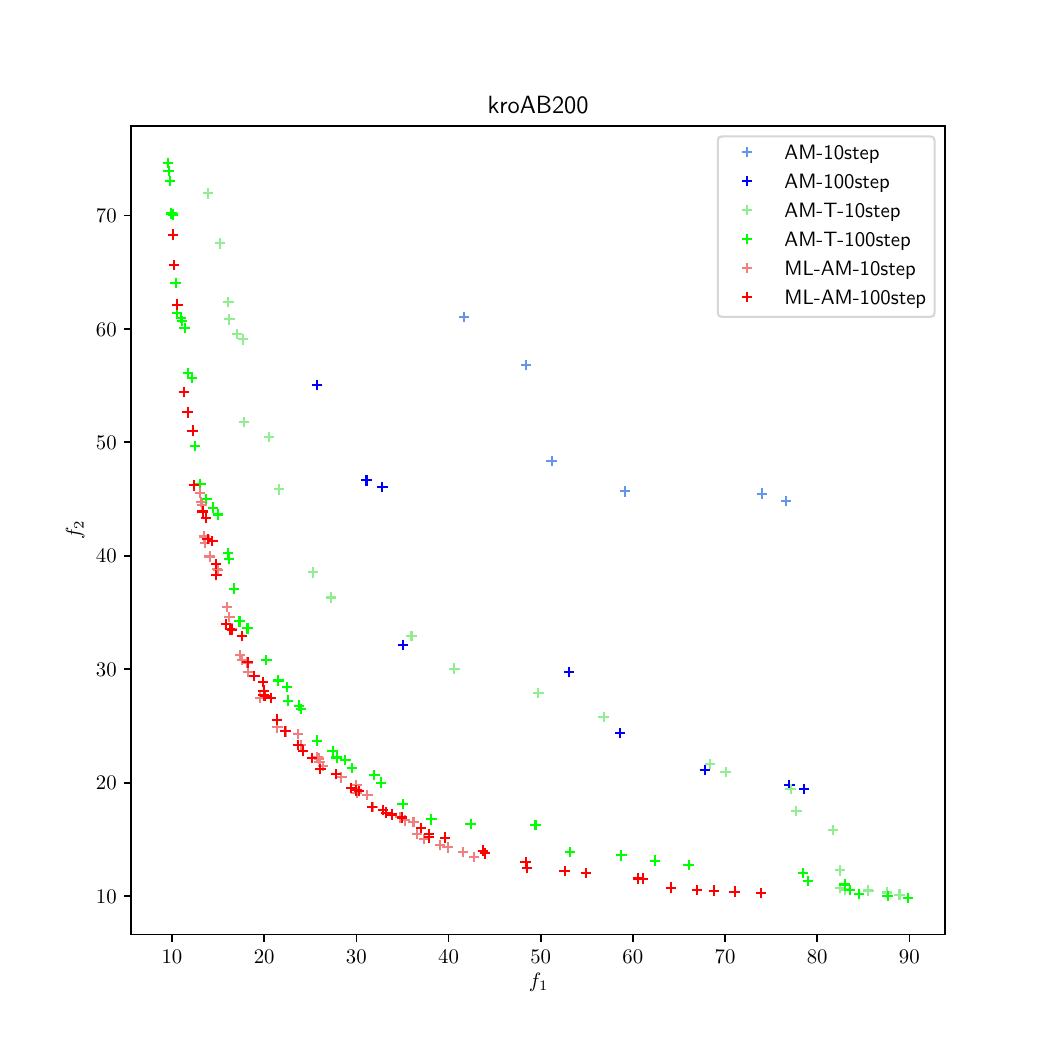}%
	}
	\caption{The PFs obtained by ML-AM, AM and AM-T on three instances (a) kroAB100, (b) kroAB150, (c) kroAB200.}
	\label{fig:kroAB-DRL}
\end{figure*}

Table~\ref{tb:kroAB-DRL} presents the summarized results of different learning-based algorithms, while \fref{fig:kroAB-DRL} plots their resultant PFs. From \fref{fig:kroAB-DRL}, we can see a clear distinction among different levels of PFs. ML-AM visibly outperforms either AM or AM-T. Table~\ref{tb:kroAB-DRL} further verifies the superiority of our method.


\begin{table}[t]
	\centering
	\caption{The results of HV and $|$NDS$|$ obtained by different learning-based algorithms on three MOTSP instances. The higher value is marked in bold.}
	\resizebox{1\columnwidth}{!}{
		\begin{tabular}{|c|c|c|c|c|c|c|c|}
			\hline
			\multirow{2}{*}{Instance} & \multirow{2}{*}{Parameter update} & \multicolumn{2}{c|}{ML-AM} & \multicolumn{2}{c|}{AM} & \multicolumn{2}{c|}{AM-T} \\ \cline{3-8} 
			&                                   & HV                 & $|$NDS$|$        & HV           & $|$NDS$|$     & HV                & $|$NDS$|$          \\ \hline
			\multirow{3}{*}{kroAB100}  & 10-step                           & \textbf{6622.20}   & \textbf{34}  & 4341.07      & 4         & 6332.42           & 27             \\ \cline{2-8} 
			& 100-step                          & \textbf{6793.54}   & \textbf{46}  & 5720.83      & 12        & 6700.96           & 37             \\ \cline{2-8} 
			& 1000-step                         & \textbf{6817.05}   & \textbf{47}  & 6434.27      & 24        & 6754.94           & 46             \\ \hline
			\multirow{3}{*}{kroAB150}  & 10-step                           & \textbf{6162.37}   & \textbf{42}  & 3037.70      & 4         & 5515.10           & 25             \\ \cline{2-8} 
			& 100-step                          & \textbf{6280.76}   & \textbf{54}  & 4746.60      & 10        & 6182.24           & 43             \\ \cline{2-8} 
			& 1000-step                         & \textbf{6280.89}   & \textbf{49}  & 5775.20      & 32        & 6272.42           & 47             \\ \hline
			\multirow{3}{*}{kroAB200}  & 10-step                           & \textbf{5603.95}   & \textbf{39}  & 1997.01      & 6         & 4555.34           & 27             \\ \cline{2-8} 
			& 100-step                          & \textbf{5796.95}   & \textbf{56}  & 3889.77      & 9         & 5630.02           & 50             \\ \cline{2-8} 
			& 1000-step                         & \textbf{5808.05}   & \textbf{59}  & 5165.18      & 32        & 5730.30           & 49             \\ \hline
		\end{tabular}
	}
	\label{tb:kroAB-DRL}
\end{table}

\begin{table*}[htbp]
	\centering
	\caption{The results of HV, $|$NDS$|$ and running time to evaluate the generalization ability of the submodels on three MOTSP instances, compared with NSGA-$\rm\uppercase\expandafter{\romannumeral2}$, MOEA/D and MOGLS. The higher HV and less time are marked in bold.}
	\resizebox{\textwidth}{!}{
		\begin{tabular}{|c|c|c|c|c|c|c|c|c|c|c|c|c|c|c|c|}
			\hline
			\multirow{2}{*}{Instance} & \multicolumn{3}{c|}{ML-AM-10step}  & \multicolumn{3}{c|}{NSGA-II-2000} & \multicolumn{3}{c|}{NSGA-II-4000} & \multicolumn{3}{c|}{MOEA/D-2000 (Tchebycheff)} & \multicolumn{3}{c|}{MOEA/D-4000 (Tchebycheff)} \\ \cline{2-16} 
			& HV               & $|$NDS$|$ & runtime(s)     & HV        & $|$NDS$|$  & runtime(s)  & HV        & $|$NDS$|$  & runtime(s)  & HV            & $|$NDS$|$       & runtime(s)      & HV            & $|$NDS$|$       & runtime(s)      \\ \hline
			kroAB100                   & \textbf{6622.20} & 34    & \textbf{29.93} & 5725.84   & 86     & 47.59       & 6104.87   & 99     & 91.22       & 5838.62       & 95          & 81.60           & 6066.24       & 99          & 161.58          \\ \hline
			kroAB150                   & \textbf{6162.37} & 42    & \textbf{42.59} & 4220.87   & 95     & 58.08       & 4508.38   & 100    & 117.30      & 4838.66       & 89          & 101.15          & 5016.97       & 92          & 184.74          \\ \hline
			kroAB200                   & \textbf{5603.95} & 39    & \textbf{59.68} & 2820.59   & 85     & 69.14       & 3501.39   & 100    & 140.04      & 3836.93       & 88          & 114.09          & 3918.80       & 87          & 217.49          \\ \hline
		\end{tabular}
	}
	\resizebox{\textwidth}{!}{
		\begin{threeparttable}
			\begin{tabular}{|c|c|c|c|c|c|c|c|c|c|c|c|c|c|c|c|}
				\hline
				\multirow{2}{*}{Instance} & \multicolumn{3}{c|}{MOEA/D-2000 (weighted sum)} & \multicolumn{3}{c|}{MOEA/D-4000 (weighted sum)} & \multicolumn{3}{c|}{MOGLS/100-1000} & \multicolumn{3}{c|}{MOGLS/100-2000} & \multicolumn{3}{c|}{MOGLS/200-1000} \\ \cline{2-16} 
				& HV            & $|$NDS$|$       & runtime(s)       & HV            & $|$NDS$|$       & runtime(s)       & HV         & $|$NDS$|$   & runtime(s)   & HV         & $|$NDS$|$   & runtime(s)   & HV         & $|$NDS$|$   & runtime(s)   \\ \hline
				kroAB100                   & 6384.70       & 98          & 76.43            & 6514.63       & 100         & 153.34           & 6382.34    & 32      & 255.49       & 6381.82    & 33      & 490.83       & 6599.46    & 41      & 455.98       \\ \hline
				kroAB150                   & 5454.63       & 89          & 88.86            & 5707.39       & 99          & 181.59           & 5436.07    & 30      & 440.99       & 5506.91    & 29      & 859.60       & 5900.62    & 52      & 805.73       \\ \hline
				kroAB200                   & 4408.21       & 89          & 102.81           & 4825.23       & 87          & 202.74           & 4500.87    & 40      & 652.29       & 4560.29    & 34      & 1298.02      & 5102.97    & 44      & 1216.66      \\ \hline
			\end{tabular}
			\begin{tablenotes}
				\footnotesize
				\item[] Note: Although $|$NDS$|$ of ML-AM is smaller than that of MOEA/D and NSGA-$\rm\uppercase\expandafter{\romannumeral2}$, the final NDS found by ML-AM completely dominates the one by others, as shown in \fref{fig:kroAB-EMO}.
			\end{tablenotes}
		\end{threeparttable}
	}
	\label{tb:kroAB-EMO}
\end{table*}

\begin{figure*}[!h]
	\centering
	\subfloat[]{\includegraphics[width=2.1in]{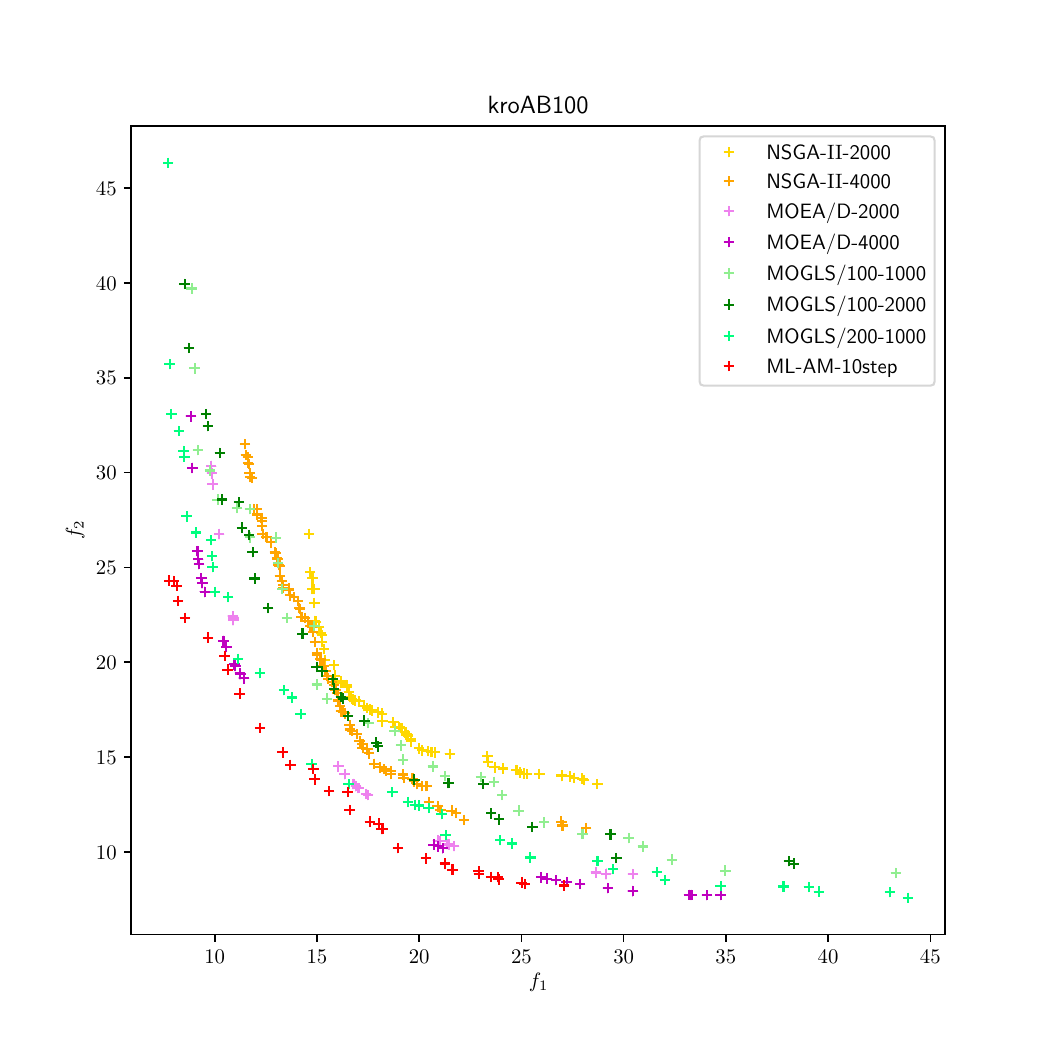}%
	}
	\hfil
	\subfloat[]{\includegraphics[width=2.1in]{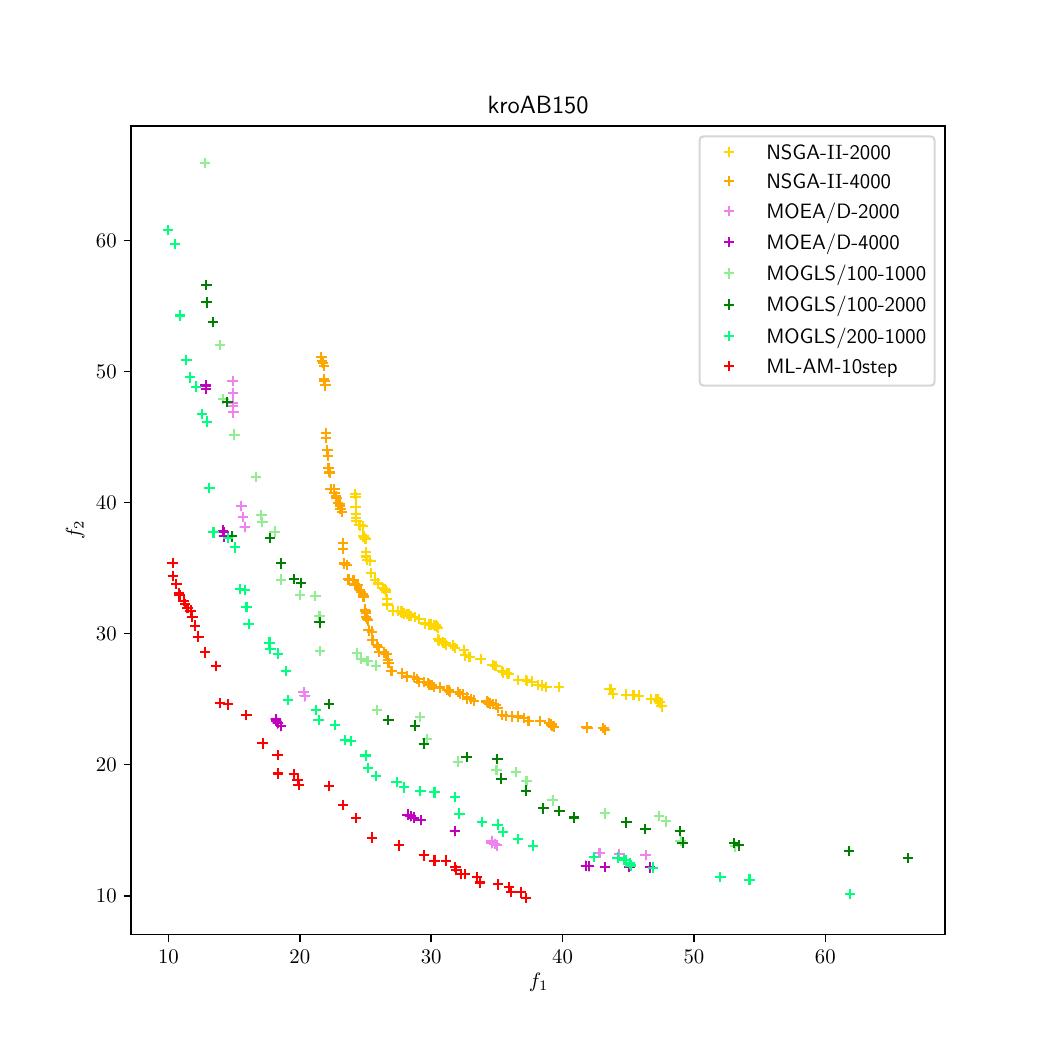}%
	}
	\hfil
	\subfloat[]{\includegraphics[width=2.1in]{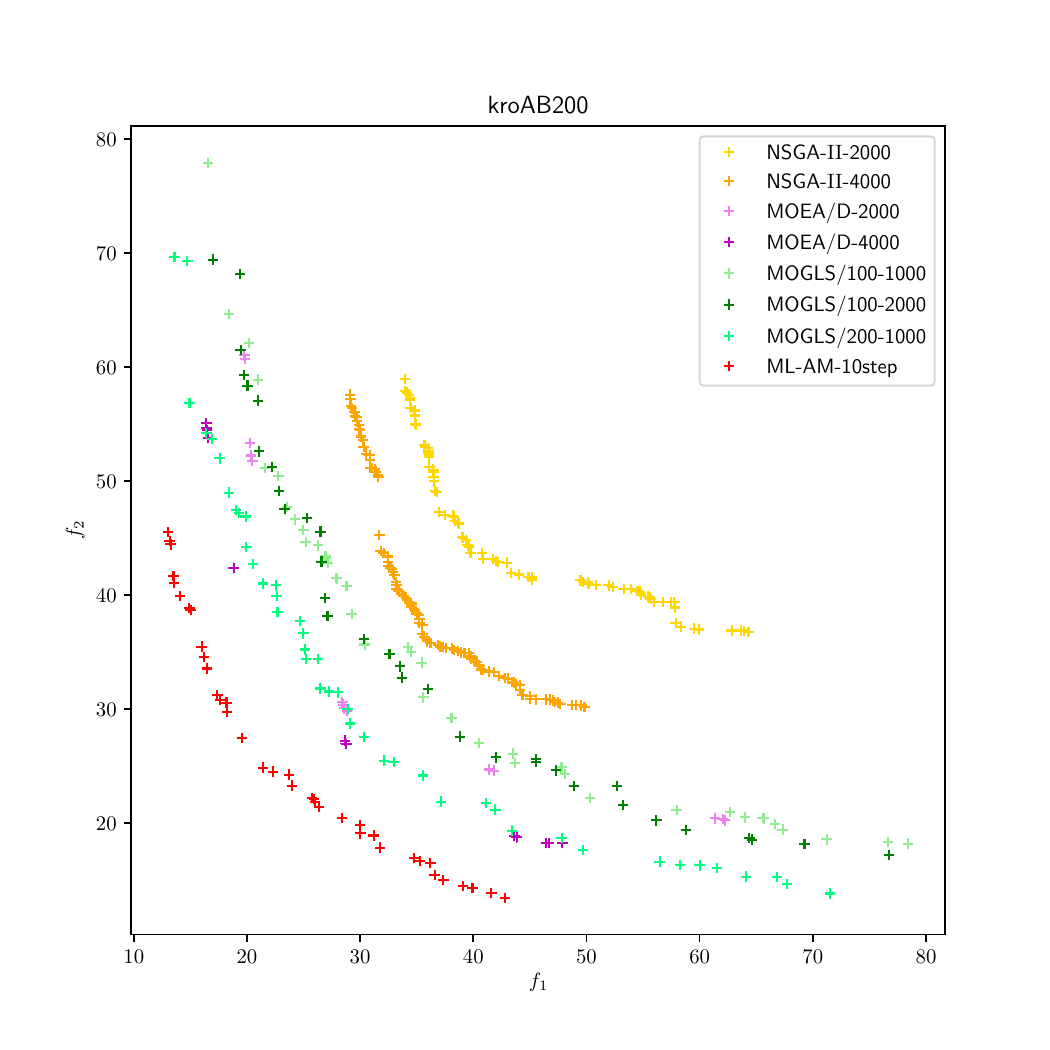}%
	}
	\caption{The PFs obtained by ML-AM, NSGA-$\rm\uppercase\expandafter{\romannumeral2}$, MOEA/D and MOGLS on three instances: (a) kroAB100, (b) kroAB150, (c) kroAB200.}
	\label{fig:kroAB-EMO}
\end{figure*}

To make more comprehensive comparisons, three iteration-based MOEAs: NSGA-$\rm\uppercase\expandafter{\romannumeral2}$, MOEA/D and MOGLS were tested. For NSGA-$\rm\uppercase\expandafter{\romannumeral2}$ and MOEA/D, the population size is set to 100. The number of iterations is set to 2000 and 4000. For MOEA/D, Tchebycheff decomposition approach and weighted sum approach are considered. For MOGLS, the maximum number of local search executions using 2-opt operations is set to either 100 or 200, while the number of iterations is set to either 1000 or 2000. We consider three combinations: MOGLS/100-1000, MOGLS/100-2000 and MOGLS/200-1000, in our experiments.

The results are reported in Table~\ref{tb:kroAB-EMO}. The corresponding PFs are shown in \fref{fig:kroAB-EMO}. The observations are as follows. 1) The submodels obtained by ML-AM (with $T$-step updates by MOTSP-50) exhibit an excellent generalization ability in coping with large-scale instances. Even with a small $T$, ML-AM is superior to most of its competitors, especially those classic MOEAs. 2) The submodels obtained by ML-AM can generate PF with good diversity. \fref{fig:kroAB-EMO} illustrates that the resultant non-dominated solutions constitute a wide spread of PF. 3) The running time of ML-AM (at the inference stage) is much faster than all the MOEAs. Note that ML-AM considers 100 subproblems and it only takes less than a second for solving a subproblem.

\section{Experiments: MOVRPTW}
\label{exp:mocvrptw}

VRPTW is much more complex than TSP, but AM can be easily modified to solve VRPTW, as described in Section \ref{sec:vrp}. Hence, the experimental settings of MLDRL on MOVRPTW are very similar to those on MOTSP.

\subsection{Experimental Settings}

\subsubsection{Test Instances}
The test instances were generated following the manner by the ``R"-group (random group) of Solomon's dataset \cite{solomon1987algorithms}. For each instance, the node features, which include the coordinates, demand and time window, were produced as follows.
\begin{itemize}
	\item The coordinates are two-dimensional vector sampled from the uniform interval of $[0, 60]\times [0, 60]$.
	\item The demand is randomly selected from the uniform integer interval of $[1, 42]$.
	\item To generate the time window $[e_i,l_i]$, the center of the time window is randomly generated between $[0, L]$, and then the width of the time window is randomly generated within the interval $[L_{min}, L_{max}]$. In our settings, $L$, $L_{min}$ and $L_{max}$ are set as 230, 10 and 200, respectively. 
	\item Once a VRPTW instance is generated, the two-dimensional coordinates are normalized to $[0,1]\times [0,1]$, and the time windows are scaled down proportionally.
\end{itemize}

\subsubsection{Other Settings}

All the experimental settings of MOVRPTW are the same as those of MOTSP, except that the meta-model is trained on randomly generated MOVRPTW instances with 50 nodes (MOVRPTW-50), and the iteration of meta-learning $T_{meta}$ is set to 2000. The reference point for calculating HV is set to $(65, 3.5)$. 

The additional difficulty of MOVRPTW compared with MOTSP is the scale imbalance issue between two objectives. For a MOVRPTW solution, its total traveling time ($f_1$) is much larger than its makespan ($f_2$). Thus, we attempt to adjust the weights to balance these two objectives \cite{zhang2007moea}. The aggregation function of the $j$-th subproblem that needs to be optimized is modified as follows:
\begin{equation}
	f^{ws}(\pi | \lambda^j)=\lambda_{1}^j \frac{f_{1}(\pi)}{f_1^*} + \lambda_{2}^j \frac{f_{2}(\pi)}{f_2^*},
	\label{eq:ws2}
\end{equation}
where $f_1^*$ and $f_2^*$ are the (approximate) global minimum value of the corresponding objectives. 

Note that the transfer learning method \cite{li2020deep} also suffers from the issue of scale imbalance. Different parameter-transfer direction may lead to different results. Therefore, we tested two parameter-transfer strategies called AM-T1 and AM-T2. Detailed descriptions of all the compared algorithms are summarized in Table \ref{tab:alg}.

\begin{table}[t]
	\centering
	\small
	\caption{Detailed descriptions of the compared algorithms for MOVRPTW.}
	\begin{tabular}{p{0.15\columnwidth}p{0.75\columnwidth}}
		\toprule
		Algorithm & Description \\
		\midrule
		ML-AM & Meta-Learning-based AM for MOVRPTW. \\
		ML-AM* & Meta-Learning-based AM with modified aggregation function for MOVRPTW. \\
		AM & AM-based DRL for each of $N=100$ VRPTW subproblems. \\
		AM* & AM-based DRL with modified aggregation function for each of $N=100$ VRPTW subproblems. \\
		AM-T1 & AM with parameter-transfer, where the initial model is trained with 5000 steps of parameter update. The transferring sequence with respect to the weights is $\lambda^1=(1,0)\rightarrow \lambda^2=(\frac{98}{99},\frac{1}{99})\rightarrow \ldots\rightarrow \lambda^{100}=(0,1)$. \\
		AM-T2 & AM with parameter-transfer, where the transferring sequence is $\lambda^1=(0,1)\rightarrow \lambda^2=(\frac{1}{99},\frac{98}{99})\rightarrow \ldots\rightarrow \lambda^{100}=(1,0)$. \\
		NSGA-II & Standard NSGA-II algorithm with the population size $N=100$ for MOVRPTW, implemented with C++. \\
		MOEA/D & Standard MOEA/D algorithm with the population size $N=100$ for MOVRPTW, implemented with C++. \\
		\bottomrule
	\end{tabular}%
	\label{tab:alg}%
\end{table}%

\begin{table*}[t]
	\centering
	\caption{The average results of HV and $|$NDS$|$ obtained by different learning-based algorithms on 128 random MOVRPTW-50 instances. The higher value is marked in bold.}
	\resizebox{\textwidth}{!}{
		\begin{threeparttable}
			\begin{tabular}{|c|c|c|c|c|c|c|c|c|c|c|c|c|c|}
				\hline
				\multirow{2}{*}{Scale} & \multirow{2}{*}{Parameter update} & \multicolumn{2}{c|}{ML-AM*} & \multicolumn{2}{c|}{ML-AM} & \multicolumn{2}{c|}{AM*} & \multicolumn{2}{c|}{AM} & \multicolumn{2}{c|}{AM-T1} & \multicolumn{2}{c|}{AM-T2} \\ \cline{3-14} 
				&                                   & HV                & $|$NDS$|$            & HV      & $|$NDS$|$          & HV                 & $|$NDS$|$              & HV                 & $|$NDS$|$             & HV                 & $|$NDS$|$              & HV                 & $|$NDS$|$      \\ \hline
				\multirow{3}{*}{50}       & 10-step                           & \textbf{80.31}    & 7.60             & 76.17           & 5.56            & 56.14    & \textbf{8.54}           & 55.46   & 8.02  & 74.50              & 5.89               & 67.40              & 7.12               \\ \cline{2-14} 
				& 100-step                          & \textbf{81.53}    & \textbf{11.09}   &75.13           & 4.90            & 61.21    & 8.78       & 62.46   & 7.51           & 79.84              & 8.33               & 75.70              & 7.44               \\ \cline{2-14} 
				& 1000-step                         & \textbf{81.62}    & \textbf{10.99}   & 81.00           & 8.83            & 75.53    & 9.70     & 74.10   & 7.33           & 81.29              & 8.98               & 79.32              & 8.59               \\ \hline
			\end{tabular}
			\begin{tablenotes}
				\footnotesize
				\item[] Note: For each step of the parameter update of a submodel, any of the six algorithms takes about 4 seconds for VRPTW-50.
			\end{tablenotes}
		\end{threeparttable}
	}
	\label{tb:MOCVRPTW-1}
\end{table*}

\begin{table*}[t]
	\centering
	\caption{The results of HV and $|$NDS$|$ obtained by different learning-based algorithms on three Solomon's instances of MOVRPTW-100. The higher value is marked in bold. }
	\resizebox{\textwidth}{!}{
		\begin{tabular}{|c|c|c|c|c|c|c|c|c|c|c|c|c|c|}
			\hline
			\multirow{2}{*}{Instance} & \multirow{2}{*}{Parameter update} & \multicolumn{2}{c|}{ML-AM*} & \multicolumn{2}{c|}{ML-AM} & \multicolumn{2}{c|}{AM*} & \multicolumn{2}{c|}{AM} & \multicolumn{2}{c|}{AM-T1} & \multicolumn{2}{c|}{AM-T2} \\ \cline{3-14} 
			&                                   & HV                  & $|$NDS$|$          & HV              & $|$NDS$|$          & HV           &$|$NDS$|$        & HV          & $|$NDS$|$      & HV                      &$|$NDS$|$        & HV                 &$|$NDS$|$             \\ \hline
			\multirow{3}{*}{R101}       & 10-step                           & \textbf{57.21}      & 3     & 53.71           & 6               & 13.42        & 4            & 12.30       & 8          & 54.68                   & 6             & 43.17              & 9                  \\ \cline{2-14} 
			& 100-step                          & \textbf{57.60}      & 8     & 54.58           & 1               & 24.85        & 8            & 27.83       & 7          & 55.78                   & 8             & 56.27              & 6                  \\ \cline{2-14} 
			& 1000-step                         & 58.56               & 8     & 53.55           & 8               & 45.24        & 10           & 44.86       & 6          & \textbf{58.88}          & 5             & 55.27              & 7                  \\ \hline
			\multirow{3}{*}{R102}       & 10-step                           & \textbf{62.89}      & 8              & 61.35           & 3               & 14.67        & 11           & 17.65       & 7          & 57.75                   & 14            & 50.11              & 9                  \\ \cline{2-14} 
			& 100-step                          & \textbf{65.83}      & 8              & 61.65           & 9               & 23.03        & 8            & 27.30       & 6          & 57.66                   & 5             & 53.15              & 9                  \\ \cline{2-14} 
			& 1000-step                         & 65.28               & 11             & 62.30           & 10              & 51.37        & 8            & 55.12       & 5          & \textbf{66.34}          & 9             & 59.93              & 6                  \\ \hline
			\multirow{3}{*}{R103}       & 10-step                           & \textbf{71.51}      & 5              & 62.50           & 6               & 11.14        & 11           & 14.56       & 5          & 61.12                   & 5             & 40.25              & 7                  \\ \cline{2-14} 
			& 100-step                          & \textbf{73.76}      & 10             & 63.65           & 6               & 18.66        & 7            & 17.51       & 6          & 65.45                   & 3             & 58.07              & 9                  \\ \cline{2-14} 
			& 1000-step                         & \textbf{74.61}      & 8              & 66.22           & 7               & 58.78        & 5            & 55.60       & 11         & 72.10                   & 10            & 64.04              & 8                  \\ \hline
		\end{tabular}
	}
	\label{tb:Solomon-DRL}
\end{table*}

\begin{table*}[t]
	\centering
	\caption{The results of HV, $|$NDS$|$ and running time to evaluate the generalization ability of the submodels on three MOVRPTW instances, compared with NSGA-$\rm\uppercase\expandafter{\romannumeral2}$ and MOEA/D. The higher HV and smaller time are marked in bold.}
	\resizebox{\textwidth}{!}{
		\begin{tabular}{|c|c|c|c|c|c|c|c|c|c|c|c|c|c|c|c|}
			\hline
			\multirow{2}{*}{Instance} & \multicolumn{3}{c|}{ML-AM*-10step} & \multicolumn{3}{c|}{NSGA-II-8000} & \multicolumn{3}{c|}{MOEA/D-8000} & \multicolumn{3}{c|}{NSGA-II-16000} & \multicolumn{3}{c|}{MOEA/D-16000} \\ \cline{2-16} 
			& HV               & $|$NDS$|$  & runtime(s)      & HV       & $|$NDS$|$   & runtime(s)   & HV      & $|$NDS$|$   & runtime(s)  & HV       & $|$NDS$|$   & runtime(s)   & HV      & $|$NDS$|$   & runtime(s)  \\ \hline
			R101                        & \textbf{57.21}   & 3      & \textbf{43.50}  & 42.47    & 5       & 97.50     &  52.93  & 7   &  170.32   
			& 52.48    &  5  &  201.34  & 55.22    &   4  &  341.98\\ \hline
			R102                        & \textbf{62.89}   & 8      & \textbf{44.91}  & 46.08    & 7       & 101.72   &  54.78  &  6  &   165.21   &
			57.51   &   4	& 203.28  & 59.58   &     6  & 335.42 \\ \hline
			R103                        & \textbf{71.51}   & 5      & \textbf{45.62}  & 46.63    & 15      & 97.26     & 59.47  &  10   &  175.74   & 52.93	 &  6  &  199.32  & 66.13  & 13  &  345.37 \\ \hline
		\end{tabular}
	}
	\label{tb:Solomon-EMO}
\end{table*}

\subsection{Experimental Results}

Table~\ref{tb:MOCVRPTW-1} shows the average performance indicators HV and $|$NDS$|$ on $128$ random instances. It can be found that $|$NDS$|$ obtained by all the learning-based methods for MOVRPTW are small. This implies that two objectives of MOVRPTW are generally consistent with each other. In this case, HV is more important to evaluate the performance of different methods.

From the table, we can see that by fine-tuning the meta-model of ML-AM* for 100 update steps, its performance is significantly better than the other methods with a thousand update steps. Even the number of update steps is set to 10, ML-AM* is able to produce satisfactory solutions. Compared ML-AM* with ML-AM, as well as AM* with AM, the results demonstrate that the modified aggregation function is mostly useful to guide the fine-tuning process towards right directions. Considering AM-T1 and AM-T2, they are all superior to AM* and AM but inferior to ML-AM* and ML-AM. AM-T1 performs better than AM-T2. The reason is that the second objective of MOVRPTW is more difficult to optimize than the first one, and the initial model of AM-T2 may not be good enough to optimize the second objective. In this case, transferring from $\lambda^1=(1,0)$ to $\lambda^{100}=(0,1)$ allows the models to have more opportunities to improve the second objective during the parameter-transfer process. In sum, ML-AM* is the most excellent approach among all the learning-based methods.

In order to test the generalization ability of the submodels fine-tuned from the meta-model in tackling different scale instances, we applied the submodels trained with MOVRPTW-50 to three commonly used Solomon's instances: R101, R102 and R103. These instances have 100 customers with their coordinates normalized into $[0,1]\times [0,1]$. The results of HV and $|$NDS$|$ for all the learning-based methods are presented in Table \ref{tb:Solomon-DRL}. The results show that ML-AM* has a good generalization ability, although it is slightly worse than AM-T1 with 1000 update steps. We can also find that the result of ML-AM* (100-step) is close to that of ML-AM* (1000-step), which means that the fine-tuned model is almost converged with only a few update steps.

Table \ref{tb:Solomon-EMO} gives the comparison results among ML-AM*, NSGA-II and MOEA/D. The numbers of iterations are set to 8000 and 16000 for both NSGA-II and MOEA/D. When we enlarge their number of iterations, the resultant HV cannot get a significant improvement. Again, ML-AM* with 10 steps of parameter update can outperform the traditional iteration-based algorithms NSGA-II and MOEA/D.

\begin{figure*}[t]
	\centering
	\subfloat[]{\includegraphics[width=2.2in]{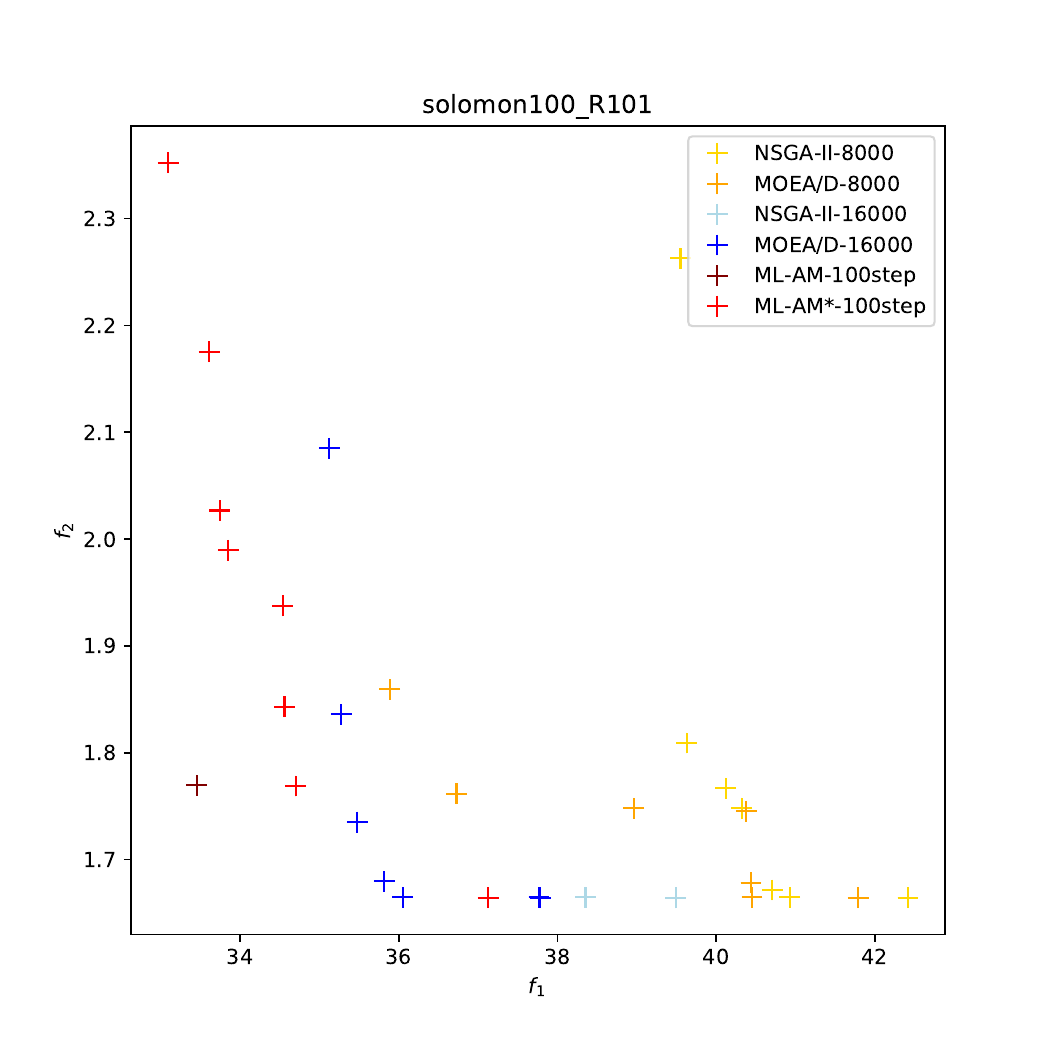}%
	}
	\hfil
	\subfloat[]{\includegraphics[width=2.2in]{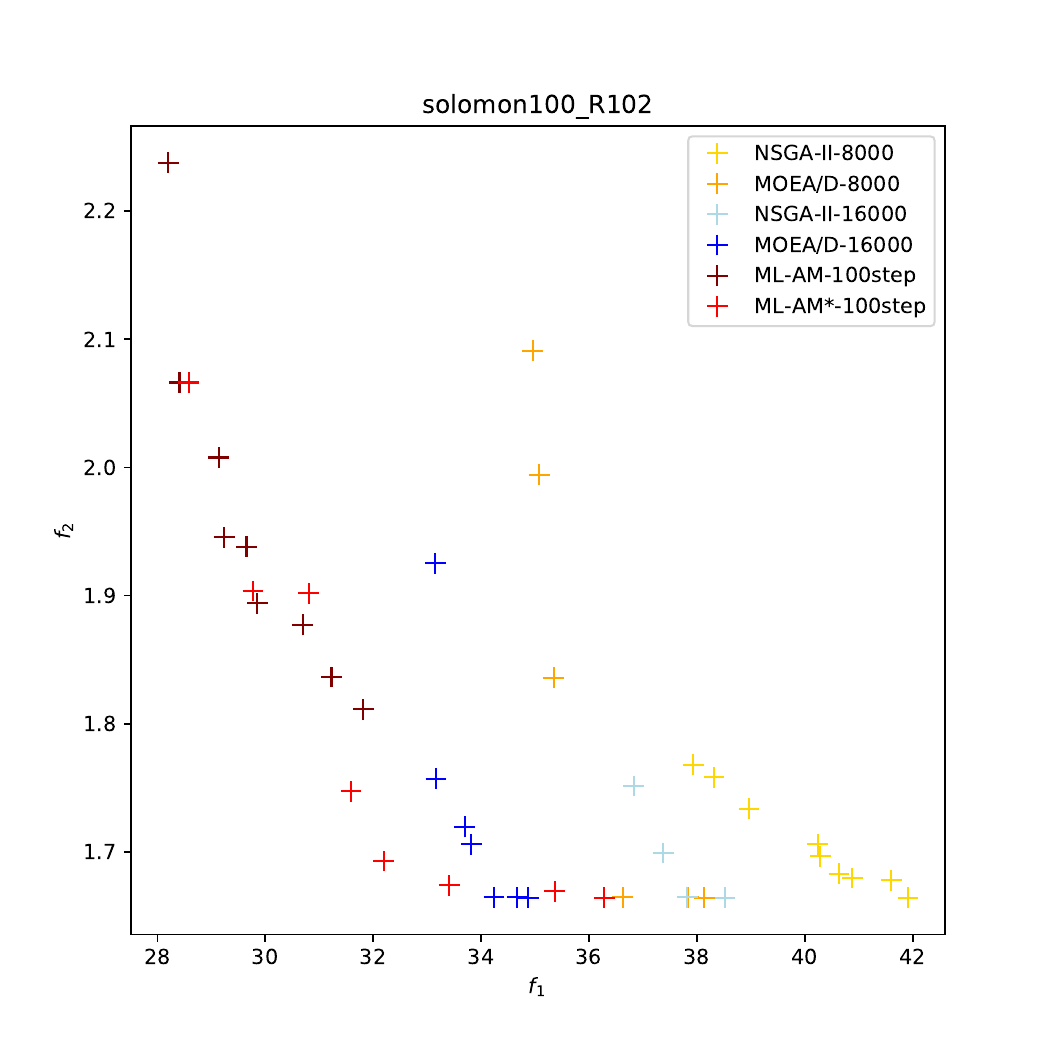}%
	}
	\hfil
	\subfloat[]{\includegraphics[width=2.2in]{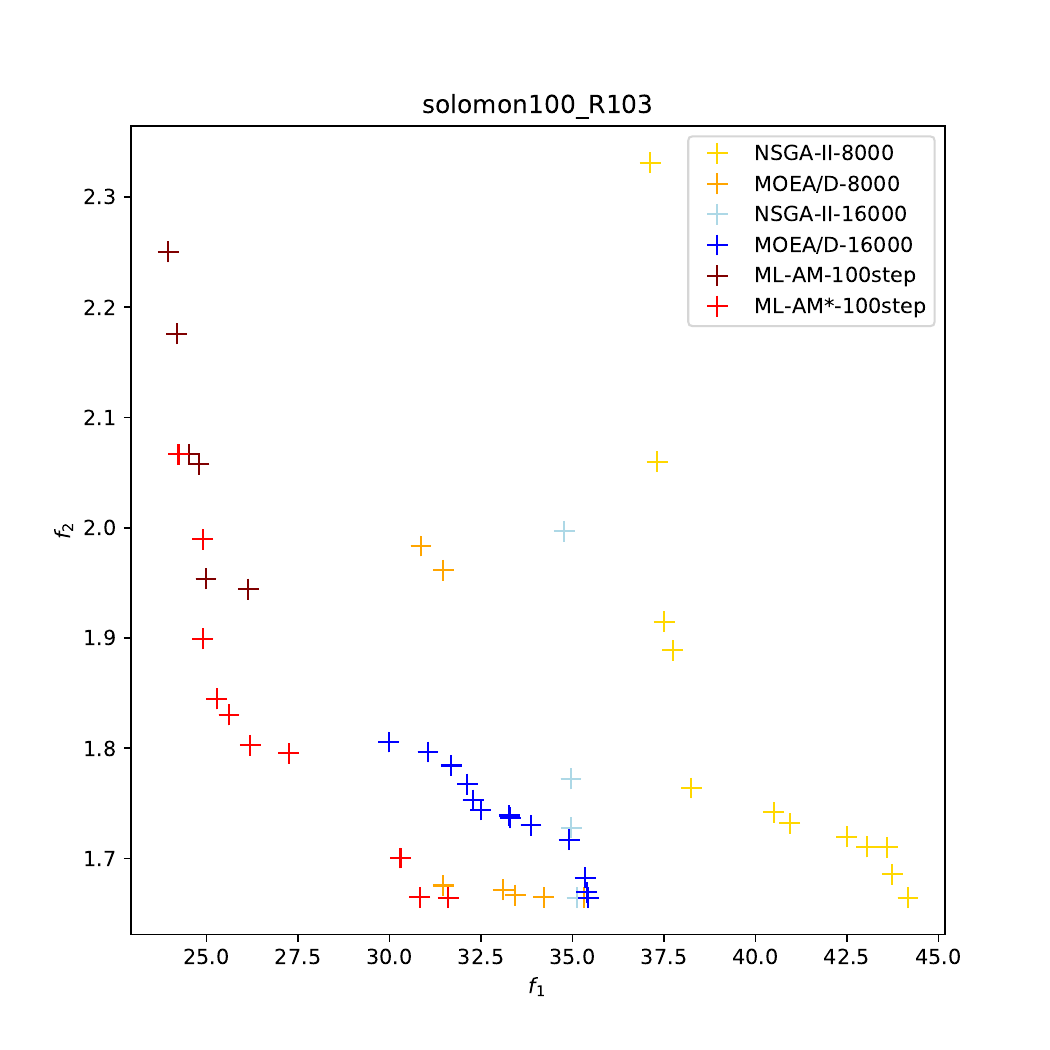}%
	}
	\caption{The PFs obtained by ML-AM*, NSGA-$\rm\uppercase\expandafter{\romannumeral2}$ and MOEA/D on three Solomon's VRPTW instances: (a) R101, (b) R102, (c) R103.}
	\label{fig:Solomon-EMO}
\end{figure*}

\fref{fig:Solomon-EMO} further visualizes the PFs obtained by three algorithms on Solomon's VRPTW instances. The non-dominated solutions found by ML-AM* locate at the lower left part of the figure and spread evenly. Different from MOTSP, MOVRPTW has a small number of non-dominated solutions found. This is because the two objectives studied (the total traveling time and makespan) has positive correlations, optimizing one objective can help to improve the other one \citep{zhang2020multi}. The figure also suggests that ML-AM* can provide a set of satisfactory trade-off solutions for the decision makers to select in the practical usage.

\section{Conclusions}
\label{6}
The recently proposed DRL methods in the field of multiobjective optimization have some shortcomings such as slow training and unable to flexibly adapt to new objective weight vectors. In this paper, we propose an MLDRL approach. It modifies the direct training process of submodels into the fine-tuning process of the meta-model, thereby greatly reducing the number of gradient update steps required to solve the subproblems, and enhancing the flexibility and versatility of the learning-based methods. In specific, we incorporate the attention model into the MLDRL framework for MOTSP and MOVRPTW. The results show that our approach not only outperforms other learning-based and iteration-based approaches, but also has excellent generalization ability in tackling the problems with different scales.

To extend our work, we plan to apply meta-learning to deal with other complex MOPs or high-dimensional MOPs (known as many-objective optimization problems). Typical problems include job shop scheduling problem \citep{zhang2020learning}, bin packing problem \citep{zhao2021online}, pickup and delivery problem \citep{li2021heterogeneous} in their multiobjective versions. It is expected that researchers would be motivated by the proposed MLDRL to devise more advanced learning-based methods for various optimization problems in the near future.

\bibliographystyle{./bibliography/IEEEtranN}

\small

\bibliography{./ref}

\newpage

\end{document}